%% file: main.tex
\documentclass{article}

\PassOptionsToPackage{numbers, compress}{natbib}

\usepackage[final]{neurips_2022}

\input{packages}
\input{macros}
\usepackage{siunitx}               
\usepackage[export]{adjustbox}

\defaultbibliography{main}
\defaultbibliographystyle{plainnat}

\title{Improving Diffusion Models for Inverse Problems using Manifold Constraints}

\author{%
Hyungjin Chung$^{*,1}$ \quad \quad Byeongsu Sim$^{*,2}$ \quad \quad Dohoon Ryu$^{1}$ \quad \quad Jong Chul Ye$^{3,1,2}$\\
  \normalfont  $^1$ Dept. of Bio and Brain Engineering\\
  $^2$ Dept. of Mathematical Sciences\\
    $^3$Kim Jaechul Graduate School of AI\\
    $^*$Equal contribution\\
  Korea Advanced Institute of Science and Technology (KAIST) \\
\texttt{\{hj.chung, byeongsu.s, dh.ryu, jong.ye\}@kaist.ac.kr} \\
}

\begin{document}

\maketitle

\begin{abstract}
Recently, diffusion models have been used to solve various  inverse problems in an unsupervised manner with appropriate modifications to the sampling process. However, the current solvers, which recursively apply a reverse diffusion step followed by a projection-based measurement consistency step, often produce sub-optimal results. By studying the generative sampling path, here we show that current solvers throw the sample path off the data manifold, and hence the error accumulates. To address this, we propose an additional correction term  inspired by the manifold constraint, which  can be used synergistically with the previous solvers to make the iterations close to the manifold. The proposed manifold constraint is straightforward to implement within a few lines of code, yet boosts the performance by a surprisingly large margin. With extensive experiments, we show that our method is superior to the previous methods both theoretically and empirically, producing promising results in many applications such as image inpainting, colorization, and sparse-view computed tomography. Code available \href{https://github.com/HJ-harry/MCG_diffusion}{here}
\end{abstract}

\section{Introduction}
\label{sec:intro}

Diffusion models have shown impressive performance both as generative models themselves~\cite{song2020score,dhariwal2021diffusion}, and also as unsupervised inverse problem solvers~\cite{kadkhodaie2020solving,choi2021ilvr,chung2022come,kawar2022denoising} that do not require problem-specific training. 
Specifically, given a pre-trained unconditional score function (i.e. denoiser), solving the reverse stochastic differential equation (SDE) numerically would amount to sampling from the data generating distribution~\cite{song2020score}. 
For many different inverse problems (e.g. super-resolution~\cite{choi2021ilvr,chung2022come}, inpainting~\cite{song2020score,chung2022come}, compressed-sensing MRI (CS-MRI)~\cite{song2022solving,chung2022come}, sparse view CT (SV-CT)~\cite{song2022solving}, etc.), it was shown that simple incorporation of the measurement process produces satisfactory conditional samples, even when the model was not trained for the specific problem.

\begin{figure}[t]
    \centering
    \includegraphics[width=1.0\textwidth]{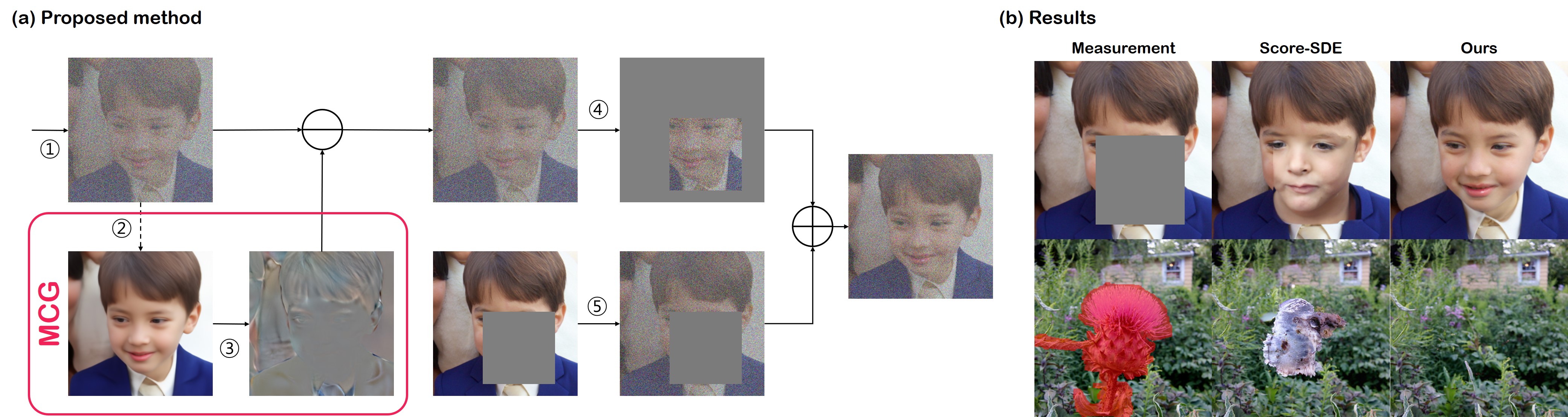}
    \caption{Visual schematic of the MCG correction step. (a)  \textcircled{\raisebox{-0.9pt}{1}} Unconditional reverse diffusion generates $\x_i$; \textcircled{\raisebox{-0.9pt}{2}} $Q_i$ maps the noisy $\x_i$ to generate $\hat{\x}_0$; \textcircled{\raisebox{-0.9pt}{3}} Manifold Constrained Gradient (MCG) $\frac{\partial}{\partial\x_i}\|\Wb(\y - \Hb\hat{\x}_0)\|_2^2$ is applied to fix the iteration on manifold; \textcircled{\raisebox{-0.9pt}{4}} Takes the orthogonal complement; \textcircled{\raisebox{-0.9pt}{5}} Samples from $p(\y_i|\y)$, then combines $\Ab\x'_{i-1}$ and $\y_i$. (b) Representative results of inpainting, compared with score-SDE~\cite{song2020score}. Reconstructions with score-SDE produce incoherent results, while our method produces high fidelity solutions.}
    \label{fig:method}  
\end{figure}

Nevertheless, for certain problems (e.g. inpainting), currently used algorithms often produce unsatisfactory results when implemented naively (e.g. boundary artifacts, as shown in Fig.~\ref{fig:method} (b)). 
The authors in \cite{lugmayr2022repaint} showed that in order to produce high quality reconstructions, one needs to iterate back and forth between the noising and the denoising step at least $> 10$ times {\em per iteration}. These iterations are computationally demanding and should be avoided, considering that diffusion models are slow to sample from even without such iterations. 
On the other hand, a classic result of Tweedie's formula~\cite{robbins1992empirical,stein1981estimation} shows that one can perform Bayes optimal denoising in one step, once we know the gradient of the log density. Extending such result, it was recently shown that one can indeed perform a single-step denoising with learned score functions for denoising problems from the general exponential family ~\cite{kim2021noisescore}.

In this work, we leverage the denoising result through Tweedie's formula and show that such denoised samples can be the key to significantly improving the performance of reconstruction using diffusion models across arbitrary linear inverse problems, despite the simplicity in the implementation.
Moreover, we theoretically prove that if the score function estimation is globally optimal, the correction term from the manifold constraint enforces the sample path to stay on the plane tangent to the data manifold\footnote{We coin our method \textbf{M}anifold \textbf{C}onstrained \textbf{G}radient (MCG).}, so by combining with the reverse diffusion step, the solution becomes more stable and accurate. 

\section{Related Works}
\label{sec:background}

\subsection{Diffusion Models}

\paragraph{Continuous Form}
For a continuous diffusion process $\x(t) \in \Rd^n,\,t \in [0, 1]$, we set $\x(0) \sim p_0(\x) = p_{data}$, where $p_{data}$ represents the data distribution of interest, and $\x(1) \sim p_1(\x)$, with $p_1(\x)$ approximating spherical Gaussian distribution, containing no information of data. 
Here, the forward noising process is defined with the following It$\hat{\text{o}}$ stochastic differential equation (SDE)~\cite{song2020score}:
\begin{equation}
\label{eq:forward-sde}
    d\x = \bar\f(\x, t)dt + \bar g(t)d\w,
\end{equation}
with $\bar\f: \Rd^d \mapsto \Rd^d$ defining the linear drift function, $\bar g(t):\Rd \mapsto \Rd$ defining a scalar diffusion coefficient, and $\w \in \Rd^n$ denoting the standard $n-$dimensional Wiener process. 
The forward SDE in \eqref{eq:forward-sde} is coupled with the following reverse SDE by the Anderson's theorem~\cite{anderson1982reverse,song2020score}:
\begin{align}\label{eq:reverse_SDE}
    d\x &= [\bar\f(\x, t) - \bar g(t)^2 \nabla_\x \log p_t(\x)]dt + \bar g(t) d\bar{\w},
\end{align}
with $dt$ denoting the infinitesimal negative time step, and $\bar{\w}$ defining the standard Wiener process running backward in time. Note that the reverse SDE defines the generative process through the score function $\nabla_\x \log~ p_t(\x)$, which in practice, is typically replaced with $\nabla_\x \log p_{0t}(\x(t)|\x(0))$
to minimize the following denoising score-matching objective
\begin{equation}
\label{eq:dsm}
    \min_\theta \Ed_{t \sim U(\varepsilon, 1), \x(0) \sim p_0(\x), \x(t) \sim p_{0t}(\x(t)|\x(0))}\left[\|\s_\theta(\x(t), t) - \nabla_{\x_t}\log p_{0t}(\x(t)|\x(0))\|_2^2\right].
\end{equation}
Once the parameter $\theta^*$ for the score function is estimated, one can replace the score function in
\eqref{eq:reverse_SDE} with $s_{\theta^*}(\x(t), t)$ to solve the reverse SDE~\cite{song2020score}.

\paragraph{Discrete Form}
Due to the linearity of $\bar\f$ and $\bar g$, the forward diffusion step can be implemented with a simple reparameterization trick~\cite{kingma2013auto}. Namely, the general form of the forward diffusion is
\begin{align}
\label{eq:forward_discrete}
    \x_i = a_i \x_0 + b_i \z,\, \quad \z \sim \Nc(0, \Ib),
\end{align}
where we have replaced the continuous index $t \in [0, 1]$ with the discrete index $i \in \Nd$.
On the other hand, the  discrete reverse diffusion step can in general be represented as
\begin{align}\label{eq:reverse_discrete}
    \x_{i-1} = \f(\x_i,\s_{\theta^*}) + g(\x_i)\z,\, \quad \z \sim \Nc(0, \Ib),
\end{align}
where we have replaced the ground truth score function with the trained one. 
We detail the choice of $a_i, b_i, \f, g$ in Appendix.~\ref{sec:app-sde}.

\subsection{Conditional Generative models for Inverse problems}

The main problem of our interest in this paper is the inverse problem, retrieving the unknown $\x \in \Rd^n$ from a measurement  $\y$:
\begin{align}\label{eq:forward}
\y = \Hb\x+\bm{\epsilon} ,\, \quad \y \in \Rd^m, \Hb \in \Rd^{m \times n},
\end{align}
where $\bm{\epsilon} \in \Rd^{m}$ is the  noise in the measurement.
Accordingly, for the case of the inverse problems,
our goal is to generate samples from a  conditional distribution with respect to the measurement $\y$, i.e. $p(\x|\y)$. Accordingly, the score function $\nabla_\x \log p_t(\x)$
in \eqref{eq:reverse_SDE} should be replaced by the conditional score $\nabla_\x \log p_t(\x|\y)$. Unfortunately, this strictly
restricts the generalization capability of the neural network since the conditional score should be retrained whenever the conditions change.
To address this, recent conditional diffusion models \cite{kadkhodaie2020solving,song2020score,choi2021ilvr,chung2022come} utilize the unconditional
score function $\nabla_\x \log p_t(\x)$ but rely on a projection-based measurement constraint to impose the conditions.
Specifically, one can apply the following:
\begin{align}
    \x'_{i-1} &= \f(\x_i,s_\theta) + g(\x_i)\z,\, \quad \z \sim \Nc(0, \Ib), \label{eq:reverse_discrete_ip}\\
    \x_{i-1} &= \Ab \x'_{i-1} + \bb_i \label{eq:nem},
\end{align}
where 
$\Ab,\bb_i$ are functions of $\Hb,\y,$ and $\x_0$.
Note that \eqref{eq:reverse_discrete_ip} is identical to the unconditional reverse diffusion step in \eqref{eq:reverse_discrete}, whereas
\eqref{eq:nem} effectively imposes the condition. Using the projection approach on Tweedie denoised estimates was proposed in~\cite{kadkhodaie2020solving}. Also, it was shown in~\cite{chung2022come} that  any general contraction mapping (e.g. projection onto convex sets, gradient step) may be utilized as \eqref{eq:nem} to impose the constraint.

Another recent work~\cite{kawar2022denoising} advancing~\cite{kawar2021snips} establishes the state-of-the-art (SOTA) in solving {\em noisy} inverse problems with unconditional diffusion models, by running the conditional reverse diffusion process in the spectral domain achieved by performing singular value decomposition (SVD), and leveraging approximate gradient of the log likelihood term in the spectral space. The authors show that feasible solutions can be obtained with as small as 20 diffusion steps.

Prior to the development of diffusion models, Plug-and-Play (PnP) models~\cite{venkatakrishnan2013plug,zhang2017learning,tirer2018image} were used in a similar fashion by utilizing a general-purpose unconditional denoiser in the place of proximal mappings in model-based iterative reconstruction methods~\cite{boyd2011distributed,beck2009fast}. 
Similarly, outside the context of diffusion models, iterative denoising followed by projection-based data consistency was proposed in~\cite{tirer2018image}. In such view, diffusion models can be understood as generative variant of PnPs trained with multiple scales of noise. 

GAN-based solvers are also widly explored~\cite{bora2017compressed,daras2021intermediate,hussein2020image}, where the pre-trained generators are tuned at the test time by optimizing over the latent, the parameters, or jointly.

\subsection{Tweedie's formula for denoising}

In the case of Gaussian noise, a classic result of Tweedie's formula~\cite{robbins1992empirical} tells us that one can achieve the denoised result by computing the posterior expectation: 
\begin{equation}
    \Ed[\x|\tilde{\x}] = \tilde\x + \sigma^2 \nabla_{\tilde\x} \log p(\tilde\x),
\end{equation} 
where the noise is modeled by $\tilde{\x} \sim \Nc(\x,\sigma^2 I)$.
If we consider a diffusion model in which the forward step is modeled as $\x_i \sim \Nc(a_i\x_0,b_i^2 I)$ (discrete form), the Tweedie's formula can be rewritten as:
\begin{equation}
\label{eq:hatx0}
    \Ed[\x_0|\x_i] = (\x_i + b_i^2 \nabla_{\x_i} \log p(\x_i))/a_i.
\end{equation}
Tweedie's formula is in fact not only relevant to Gaussian denoising in the Bayesian framework, but have also been extended to be in close relation with kernel regression~\cite{ong2019local}. Moreover, it was shown that it can be applied to arbitrary exponential noise distributions beyond Gaussian~\cite{efron2011tweedie,kim2021noisescore}. In the following, we use this key property to develop our algorithm.

\section{Conditional Diffusion using Manifold Constraints}

Although our original motivation of using the measurement constraint step in \eqref{eq:nem} was to utilize the unconditionally trained score function in  the reverse diffusion step in \eqref{eq:reverse_discrete_ip}, there is room for imposing additional constraints while still using the unconditionally trained score function.

Specifically, the Bayes rule $p(\x|\y)=p(\y|\x)p(\x)/p(\y)$ 
leads to 
\begin{align}
\label{eq:cond_score}
\nabla_\x \log p(\x|\y) = \nabla_{\x} \log p(\x)+ \nabla_\x\log p(\y|\x).
\end{align}
Hence,  the score function in the reverse SDE in \eqref{eq:reverse_discrete_ip} can be replaced by \eqref{eq:cond_score}, leading to
\begin{align}
\label{eq:notweedie}
    \x'_{i-1} &= \f(\x_i,\s_\theta) - \alpha\frac{\partial}{\partial\x_i} \|{\boldsymbol W}({\y} - \Hb\x_i)\|_2^2 + g(\x_i)\z,\,\quad  \z \sim \Nc(0, \Ib) 
\end{align}
where $\alpha$ and $\Wb$ depend on the noise covariance, if the noise $\bm{\epsilon}$ in \eqref{eq:forward} is Gaussian.

Now, one of the important contributions of this paper is to reveal that the Bayes optimal denoising step in \eqref{eq:hatx0} from the Tweedie's formula leads to a preferred condition  both empirically and theoretically. 
Specifically, we define the set constraint  for $\x_i$, called the {\em manifold constrained gradient (MCG)}, so that the gradient of the measurement term stays on the manifold (see Theorem~\ref{thm:MCG}):
\begin{align}
\x \in \Xc_i,\quad&\mbox{where}\quad \Xc_i = \{\x \in \Rd^n~|~\x = ( \x+b_i^2 \s_\theta(\x,i))/a_i \}
\end{align}
To deal with the potential deviation from the measurement consistency, we again impose the data consistency step \eqref{eq:nem}.
Putting them together, the discrete reverse diffusion under the additional manifold constraint and the data consistency can be represented by
\begin{align}
    \x'_{i-1} &= \f(\x_i,\s_\theta) - \alpha\frac{\partial}{\partial\x_i} \|{\boldsymbol W}(\y - \Hb\hat\x_0(\x_i))\|_2^2 + g(\x_i)\z,\,\quad  \z \sim \Nc(0, \Ib), \label{eq:reverse_discrete_ip_mcg}\\
    \x_{i-1} &= \Ab \x'_{i-1} + \bb \label{eq:nem_mcg}. 
\end{align}
We illustrate our scheme visually in Fig.~\ref{fig:method} (a), specifically for the task of image inpainting. The additional step leads to a dramatic performance boost, as can be seen in Fig.~\ref{fig:method} (b). Note that while the mapping \eqref{eq:hatx0} does not rely on the measurement, our gradient term in \eqref{eq:reverse_discrete_ip_mcg} incorporates the information of $\y$ so that
the gradient of the measurement terms stays on the manifold.
In the following, we study the theoretical properties of the method. Further algorithmic details and adaptations to each problem that we tackle are presented in Section~\ref{sec:app-algo}.

We note that the authors of~\cite{ho2022video} proposed a similar gradient method for the application of temporal imputation and super-resolution. When combining \eqref{eq:reverse_discrete_ip_mcg} with \eqref{eq:nem_mcg}, one can arrive at a similar gradient method proposed in~\cite{ho2022video}, and hence our method can be seen as a generalization to arbitrary linear inverse problems.
Furthermore, there are vast literature in the context of PnP models that utilize pre-trained denoisers together with gradient of the log-likelihood to solve inverse problems~\cite{laumont2022bayesian,vidal2020maximum,de2020maximum}. Among them, \cite{laumont2022bayesian} is especially relevant to this work since their method relies on modified Langevin diffusion, together with Tweedie's denoising and projections to the measurement subspace.

\section{Geometry of Diffusion Models and Manifold Constrained Gradient}
\label{sec:theory}

\begin{figure*}[t!]
    \centering
    \begin{subfigure}[b]{0.4\textwidth}
        \centering
        \includegraphics[width=\textwidth]{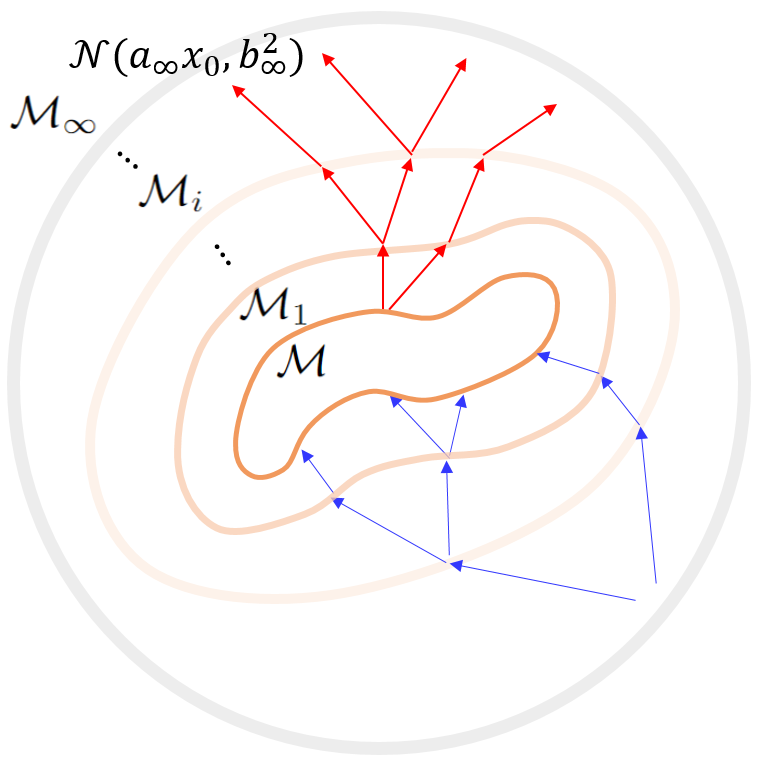}
        \caption{Geometry of diffusion model}
        \label{fig:theory geometry}
    \end{subfigure}
    \hfill
    \begin{subfigure}[b]{0.5\textwidth}
        \centering
        \includegraphics[width=\textwidth,height=4.5cm]{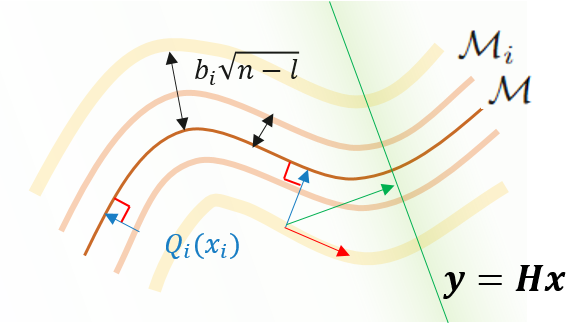}
        \caption{MCG correction}
        \label{fig:theory MCG}
    \end{subfigure}
	\caption{In both (a) and (b), the central manifolds represent the data manifold $\Mc$, encircled by manifolds of noisy data $\Mc_i$. The concentration on the manifold of noisy data and the distance from the clean data manifold are prescribed by Proposition~\ref{prop:noisy}. In (a), the backward (resp. forward) step depicted by blue (resp. red) arrows can be considered as transitions from $\Mc_i$ to $\Mc_{i-1}$ (resp. $\Mc_{i-1}$ to $\Mc_i$). In (b), arrows refer to the directions of conventional projection onto convex sets (POCS) step (green arrow) and MCG step (red arrow) which can be predicted by Theorem \ref{thm:MCG}.}
	\label{fig:theory}
\end{figure*}

In this section, we theoretically support the effectiveness of the proposed algorithm by showing the problematic behavior of the earlier algorithm and how the proposed algorithm resolves the problem.
We defer all proofs in the supplementary section.
To begin with, we borrow a geometrical viewpoint of the data manifold.

\textbf{Notation} For a scalar $a$, points $\x,\y$ and a set $A$, we use the following notations.
$aA:= \{a \x: \x \in A\}$;
$d(\x,A) := \inf_{\y \in A} ||\x-\y||_2$;
$B_r(A):= \{\x: d(\x, A) < r\}$; $T_\x \Mc$: the tangent space to a manifold $\Mc$ at $\x$; $\Jb_f$: the Jacobian matrix of a vector valued function $f$. We define $p_0 = p_{data}$.

To develop the theory, we need an assumption on the data distribution.
\begin{restatable}[Strong manifold assumption: linear structure]{assumption}{manifoldassumption}
    Suppose $\Mc\subset \Rd^n$ is the set of all data points, here we call the data manifold. Then, the manifold coincides with the tangent space with dimension $l\ll n$.
    $$\mathcal{M} \cap B_R(\x_0) = T_{x_0} \Mc \cap B_R(x_0) \text{ and } T_{x_0} \Mc \cong \Rd^l.$$
    Moreover, the data distribution $p_0$ is the uniform distribution on the data manifold $\Mc$.
\end{restatable}
We need to recall that the conventional manifold assumption is about the intrinsic geometry of data points having a low dimensional nature. However, we assume more in this work: the manifold is locally linear. Although this stronger assumption might narrow the practice of the theory, the geometric approach may provide new insights on diffusion models.
Under this assumption, the following proposition shows how the data perturbed by noise lies in the ambient space, illustrated pictorially in  Fig.~\ref{fig:theory geometry}.
\begin{restatable}[Concentration of noisy data]{proposition}{noisymanifold}
    \label{prop:noisy}
    Consider the distribution of noisy data $p_i(\x_i) = \int p(\x_i|\x)p_0(\x) d\x, p(\x_i|\x) \sim \Nc(a_i\x, b_i^2 \Ib) $.
    Then $p_i(\x_i)$ is concentrated on $(n-1)$-dim manifold $\Mc_i := \{ \y \in \Rd^n : d(\y, a_i \Mc) = r_i := b_i \sqrt{n-l} \}$. Rigorously, $p_i(B_{\epsilon r_i}(\Mc_i)) > 1 - \delta$, for some small $\epsilon,\delta>0$.
\end{restatable}
\begin{remark}[Geometric interpretation of the diffusion process]
Considering  Proposition~\ref{prop:noisy}, the manifolds of noisy data can be interpreted as interpolating manifolds between the two: the hypersphere, where pure noise $\Nc(a_\infty\x_0,b_\infty^2)$ is concentrated, and the clean data manifold.
In this regard, the diffusion steps are mere transitions from one manifold to another and the diffusion process is a transport from the data manifold to the hypersphere through interpolating manifolds. See Fig.~\ref{fig:theory geometry}.
\end{remark}
\begin{remark}
\label{rmk:score}
We can infer from the proposition that the score functions are trained only with the data points concentrated on the noisy data manifolds. Therefore, inaccurate inference might be caused by application of a score function on points away from the noisy data manifold.
\end{remark}

\begin{restatable}[score function]{proposition}{score}
\label{prop:score}
Suppose $s_\theta$ is the minimizer of the denoising score matching loss in \eqref{eq:dsm}.
Let $Q_i$ be the function that maps $\x_i$ to $\hat{\x}_0$ for each $i$, $$Q_i:\Rd^d \rightarrow \Rd^d, \x_i \mapsto \hat{\x}_0:=\frac{1}{a_i} (\x_i + b_i^2 s_\theta (\x_i,i) ).$$ 
Then, $Q_i(\x_i) \in \Mc$ and $\Jb_{Q_i}^2 = \Jb_{Q_i} = \Jb_{Q_i}^T: \Rd^d \rightarrow T_{Q_i(\x_i)}\Mc$. Intuitively, $Q_i$ is locally an orthogonal projection onto $\Mc$.
\end{restatable}

According to the proposition, the score function only concerns the normal direction of the data manifold.
In other words, the score function cannot discriminate  two data points whose difference is tangent to the manifold.
In solving inverse problems, however, we desire to discriminate data points to reconstruct the original signal, and the discrimination is achievable by measurement fidelity.
In order to achieve the original signal, the measurement plays a role in correcting the tangent component near the data manifold.
Furthermore, with regard to \cref{rmk:score}, diffusion model-based inverse problem solvers should follow the tangent component.
The following theorem shows how existing algorithms and the proposed method are different in this regard.

\begin{restatable}[Manifold constrained gradient]{theorem}{Improved}
A correction by the manifold constrained gradient does not leave the data manifold. Formally,
    \begin{align*}
        \frac{\partial}{\partial\x_i} \|\Wb(\y - \Hb\hat\x_0)\|_2^2 =-2 \Jb_{Q_i}^T \Hb^T \Wb^T \Wb (\y - \Hb\hat\x_0) \in T_{\hat\x_0}\Mc,
    \end{align*}
the gradient is the projection of the data fidelity term onto $T_{\hat\x_0}\Mc$,
\label{thm:MCG}
\end{restatable}
This theorem suggests that in diffusion models, the naive measurement fidelity step (without considering the data manifold) pushes the inference path out of the manifolds and might lead to inaccurate reconstruction. (To see this pictorially, see section.~\ref{sec:evolution}, and Fig.~\ref{fig:evolution}.)
On the other hand, our correction term from the manifold constraint guides the diffusion to lie on the data manifold, leading to better reconstruction.
Such geometric views are illustrated in Fig.~\ref{fig:theory MCG}. 
\begin{remark}
One may concern that the suboptimality of the denoising score matching loss optimization may lead to inaccurate inference of the MCG steps.
In practice, however, most of the error in denoising score matching is concentrated on $t\sim 1$\cite{chung2022come}, and in such region, the Tweedie's inference cannot make meaningful images. That is, the score function cannot detect the data manifold.
Nonetheless, in this regime, the magnitudes of the MCGs are small when the denoising score is inaccurate, and hence the matters arising from suboptimality is minimal. As $t \rightarrow 0$, the estimation becomes exact, and subsequently leads to accurate implementation of the MCG.
\end{remark}

\section{Experiments}
\label{sec:exp}

 \begin{figure}[t]
    \centering
    \includegraphics[width=0.9\textwidth]{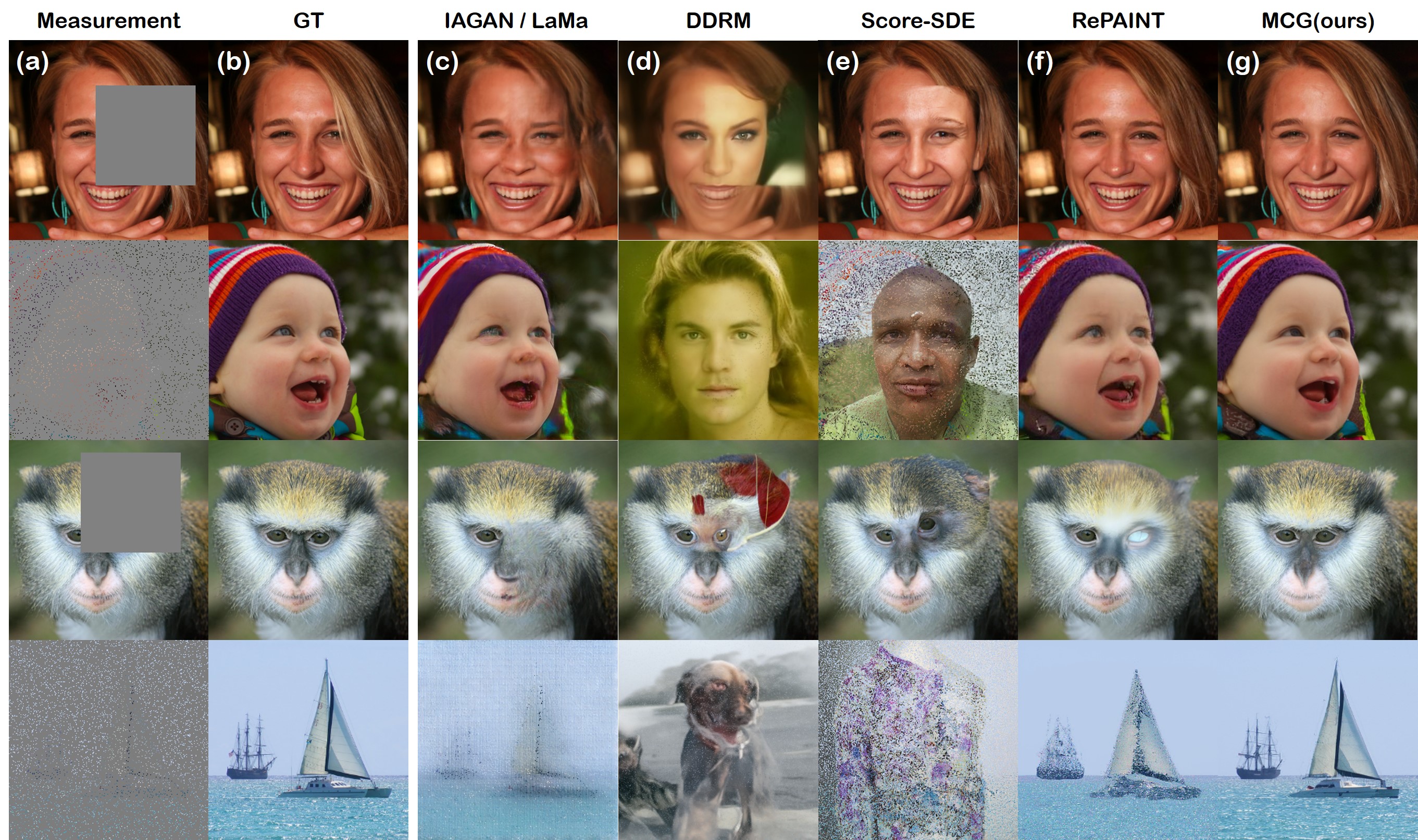}
    \caption{Inpainting results on FFHQ (1st, 2nd row) and ImageNet (3rd, 4th row). (a) Measurement, (b) Ground truth, (c) IAGAN~\cite{hussein2020image} for FFHQ, LaMa~\cite{suvorov2022resolution} for ImageNet, (d) DDRM~\cite{kawar2022denoising}, (e) Score-SDE~\cite{song2020score}, (f) RePAINT~\cite{lugmayr2022repaint}, (g) MCG (Ours). Out of 256 $\times$ 256 image, the 1st and the 3rd row is masked with size $128 \times 128$ box. 92\% of pixels (all RGB channels) from the images in the 2nd and 4th row are blocked.}
    \label{fig:results_inpainting}
\end{figure}

For all tasks, we aim to verify the superiority of our method against other diffusion model-based approaches, and also against strong supervised learning-based baselines. Further details can be found in Section.~\ref{sec:app_exp_detail}.

\paragraph{Datasets and Implementation}

For inpainting, we use FFHQ 256$\times$256~\cite{karras2019style}, and ImageNet 256$\times$256~\cite{deng2009imagenet} to validate our method. We utilize pre-trained models from the open sourced repository based on the implementation of ADM (VP-SDE)~\cite{dhariwal2021diffusion}. We validate the performance on 1000 held-out validation set images for both FFHQ and ImageNet dataset.
For the colorization task, we use FFHQ 256$\times$256, and LSUN-bedroom 256$\times$256~\cite{yu2015lsun}. We use pre-trained score functions from score-SDE~\cite{song2020score} based on VE-SDE. We use 300 validation images for testing the performance with respect to the LSUN-bedroom dataset.
For experiments with CT, we train our model based on \texttt{ncsnpp} as a VE-SDE from score-SDE~\cite{song2020score}, on the 2016 American Association of Physicists in Medicine (AAPM) grand challenge dataset, and we process the data as in~\cite{kang2017deep}. Specifically, the dataset contains 3839 training images resized to 256$\times$256 resolution. We simulate the CT measurement process with parallel beam geometry with evenly-spaced 180 degrees. Evaluation is performed on 421 held-out validation images from the AAPM challenge.

\begin{table*}[]
\centering
\setlength{\tabcolsep}{0.2em}
\resizebox{1.0\textwidth}{!}{
\begin{tabular}{lllllllll@{\hskip 15pt}llllll}
\toprule
{} & \multicolumn{8}{c}{\textbf{FFHQ} ($\bf 256 \times 256$)} & \multicolumn{6}{c}{\textbf{ImageNet} ($\bf 256 \times 256$)} \\
\cmidrule(lr){2-9}
\cmidrule(lr){10-15}
{} & \multicolumn{2}{c}{\textbf{Box}} & \multicolumn{2}{c}{\textbf{Random}} & \multicolumn{2}{c}{\textbf{Extreme}} &
\multicolumn{2}{c}{\textbf{Wide masks}} &\multicolumn{2}{c}{\textbf{Box}} & \multicolumn{2}{c}{\textbf{Random}} & \multicolumn{2}{c}{\textbf{Wide masks}} \\
\cmidrule(lr){2-3}
\cmidrule(lr){4-5}
\cmidrule(lr){6-7}
\cmidrule(lr){8-9}
\cmidrule(lr){10-11}
\cmidrule(lr){12-13}
\cmidrule(lr){14-15}
{\textbf{Method}} & {FID $\downarrow$} & {LPIPS $\downarrow$} & {FID $\downarrow$} & {LPIPS $\downarrow$} & {FID $\downarrow$} & {LPIPS $\downarrow$} & {FID $\downarrow$} & {LPIPS $\downarrow$} & {FID $\downarrow$} & {LPIPS $\downarrow$} & {FID $\downarrow$} & {LPIPS $\downarrow$} & {FID $\downarrow$} & {LPIPS $\downarrow$} \\
\midrule
MCG~\textcolor{trolleygrey}{(ours)} & \textbf{23.7} & \underline{0.089} & \textbf{21.4} & \textbf{0.186} & \textbf{30.6} & \textbf{0.366} & \textbf{22.1} & \underline{0.099} &\textbf{25.4} & 0.157 & \textbf{34.8} & \textbf{0.308} & \underline{21.9} & \underline{0.148}\\
\cmidrule(l){1-15}
Score-SDE~\cite{song2020score} & 30.3 & 0.135 & 109.3 & 0.674 & 48.6 & 0.488 & 29.8 & 0.132 & 43.5 & 0.199 & 143.5 & 0.758 & 25.9 & 0.150\\
RePAINT$^*$~\cite{lugmayr2022repaint} & \underline{25.7} & 0.093 & \underline{38.1} & \underline{0.240} & \underline{35.9} & \underline{0.398} & 24.2 & 0.108 & \underline{26.1} & 0.156 & \underline{59.3} & \underline{0.387} & 37.0 & 0.205\\
DDRM~\cite{kawar2022denoising} & 28.4 & 0.109 & 111.6 & 0.774 & 48.1 & 0.532 & 27.5 & 0.113 & 88.8 & 0.386 & 99.6 & 0.767 & 80.6 & 0.398\\
LaMa~\cite{suvorov2022resolution} & 27.7 & \textbf{0.086} & 188.7 & 0.648 & 61.7 & 0.492 & \underline{23.2} & \textbf{0.096} & 26.8 & \textbf{0.139} & 134.1 & 0.567 & \textbf{20.4} & \textbf{0.140}\\
AOT-GAN~\cite{zeng2022aggregated} & 29.2 & 0.108 & 97.2 & 0.514 & 69.5 & 0.452 & 28.3 & 0.106 & 35.3 & 0.163 & 119.6 & 0.583 & 29.8 & 0.161 \\
ICT~\cite{wan2021high} & 27.3 & 0.103 & 91.3 & 0.445 & 56.7 & 0.425 & 26.9 & 0.104 & 31.9 & \underline{0.148} & 131.4 & 0.584 & 25.4 & 0.148\\
DSI~\cite{peng2021generating} & 27.9 & 0.096 & 126.4 & 0.601 & 77.5 & 0.463 & 28.3 & 0.102 & 34.5 & 0.155 & 132.9 & 0.549 & 24.3 & 0.154\\
IAGAN~\cite{hussein2020image} & 26.3 & 0.098 & 41.5 & 0.279 & 56.1 & 0.417 & 23.8 & 0.110 & - & - & - & - & - & - \\
\bottomrule
\end{tabular}
}
\vspace{0.2em}
\caption{
Quantitative evaluation (FID, LPIPS) of inpainting task on FFHQ and ImageNet. $^*$: Re-implemented with our score function. MCG, Score-SDE, RePAINT, and DDRM all share the same score function and differ only in the inference method. \textbf{Bold}: Best, \underline{under}: second best.
}
\label{tab:comparison-inpainting}
\end{table*}

\paragraph{Inpainting}

\begin{wraptable}[11]{r}{0.5\textwidth}
\centering
\setlength{\tabcolsep}{0.2em}
\resizebox{0.4\textwidth}{!}{
\begin{tabular}{lllll}
\toprule
\cmidrule(lr){2-5}
{Data} & \multicolumn{2}{c}{\textbf{FFHQ(256$\times$256)}} & \multicolumn{2}{c}{\textbf{LSUN(256$\times$256)}} \\
\cmidrule(lr){2-3}
\cmidrule(lr){4-5}
{\textbf{Method}} & {SSIM $\uparrow$} & {LPIPS $\downarrow$} & {SSIM $\downarrow$} & {LPIPS $\downarrow$} \\
\midrule
MCG~\textcolor{trolleygrey}{(ours)} & \underline{0.951} & \textbf{0.146} & \textbf{0.959} & \textbf{0.160}\\
\cmidrule(l){1-5}
Score-SDE~\cite{song2020score} & 0.936 & 0.180 & 0.945 & 0.199\\
DDRM~\cite{kawar2022denoising} & 0.948 & \underline{0.154} & \underline{0.957} & \underline{0.182}\\
cINN~\cite{ardizzone2019guided} & \textbf{0.952} & 0.166 & 0.952 & 0.180 \\
pix2pix~\cite{isola2017image} & 0.935 & 0.184 & 0.947 & 0.174 \\
\bottomrule
\end{tabular}
}
\vspace{0.2em}
\caption{
Quantitative evaluation (SSIM, LPIPS) of colorization task. \textbf{Bold}: best, \underline{under}: second best.
}
\label{tab:comparison-color}
\end{wraptable}

Score-SDE~\cite{song2020score}, REPAINT~\cite{lugmayr2022repaint}, DDRM~\cite{kawar2022denoising} were chosen as baseline diffusion models to compare against the proposed method. For a fair comparison, we use the same score function for all methods including MCG, and only differentiate the inference method that is used. Another class of generative models: GAN-based inverse problem solver, IAGAN~\cite{hussein2020image} is considered as a comparison method for FFHQ specifically.
We also include comparisons against supervised learning based baselines: LaMa~\cite{suvorov2022resolution}, AOT-GAN~\cite{zeng2022aggregated}, ICT~\cite{wan2021high}, and DSI~\cite{peng2021generating}. We use various forms of inpainting masks: box (128 $\times$ 128 sized square region is missing\footnote{The location of the box is sampled uniformly within 16 pixel margin of each side.}), extreme (only the box region is existent), random (90-95\% of pixels are missing), and LaMa-wide. Quantitative evaluation is performed with two metrics - Frechet Inception Distance (FID)-1k~\cite{NIPS2017_8a1d6947}, and Learned Perceptual Image Patch Similarity (LPIPS)~\cite{zhang2018unreasonable}.

\begin{wraptable}[11]{r}{0.5\textwidth}
\centering
\setlength{\tabcolsep}{0.2em}
\resizebox{0.4\textwidth}{!}{
\begin{tabular}{lllll}
\toprule
{} & \multicolumn{4}{c}{\textbf{AAPM} ($\bf 256 \times 256$)}\\
\cmidrule(lr){2-5}
{Views} & \multicolumn{2}{c}{\textbf{18}} & \multicolumn{2}{c}{\textbf{30}} \\
\cmidrule(lr){2-3}
\cmidrule(lr){4-5}
{\textbf{Method}} & {PSNR $\uparrow$} & {SSIM $\uparrow$} & {PSNR $\uparrow$} & {SSIM $\uparrow$} \\
\midrule
MCG~\textcolor{trolleygrey}{(ours)} & \textbf{33.57} & \textbf{0.956} & \textbf{36.09} & \textbf{0.971}\\
\cmidrule(l){1-5}
Score-CT~\cite{song2022solving} & 29.85 & 0.897 & 31.97 & 0.913\\
SIN-4c-PRN~\cite{wei20202} & 26.96 & 0.850 & 30.23 & 0.917 \\
cGAN~\cite{ghani2018deep} & 24.38 & 0.823 & 27.45 & 0.927 \\
FISTA-TV~\cite{beck2009fast} & 21.57 & 0.791 & 23.92 & 0.861\\
\bottomrule
\end{tabular}
}
\vspace{0.2em}
\caption{
Quantitative evaluation (PSNR, SSIM) of CT reconstruction task. \textbf{Bold}: best.
}
\label{tab:comparison-ct}
\end{wraptable}

Our method outperforms the diffusion model baselines~\cite{song2020score,lugmayr2022repaint,kawar2022denoising} by a large margin. Moreover, our method is also competitive with, or even better than the best-in-class fully supervised methods, as can be seen in Table~\ref{tab:comparison-inpainting}. In Fig.~\ref{fig:results_inpainting}, we depict representative results that show the superiority of the method, where we see in both the box-type and random dropping that MCG performs very well on all experiments.

\begin{figure}[t]
    \centering
    \includegraphics[width=\textwidth]{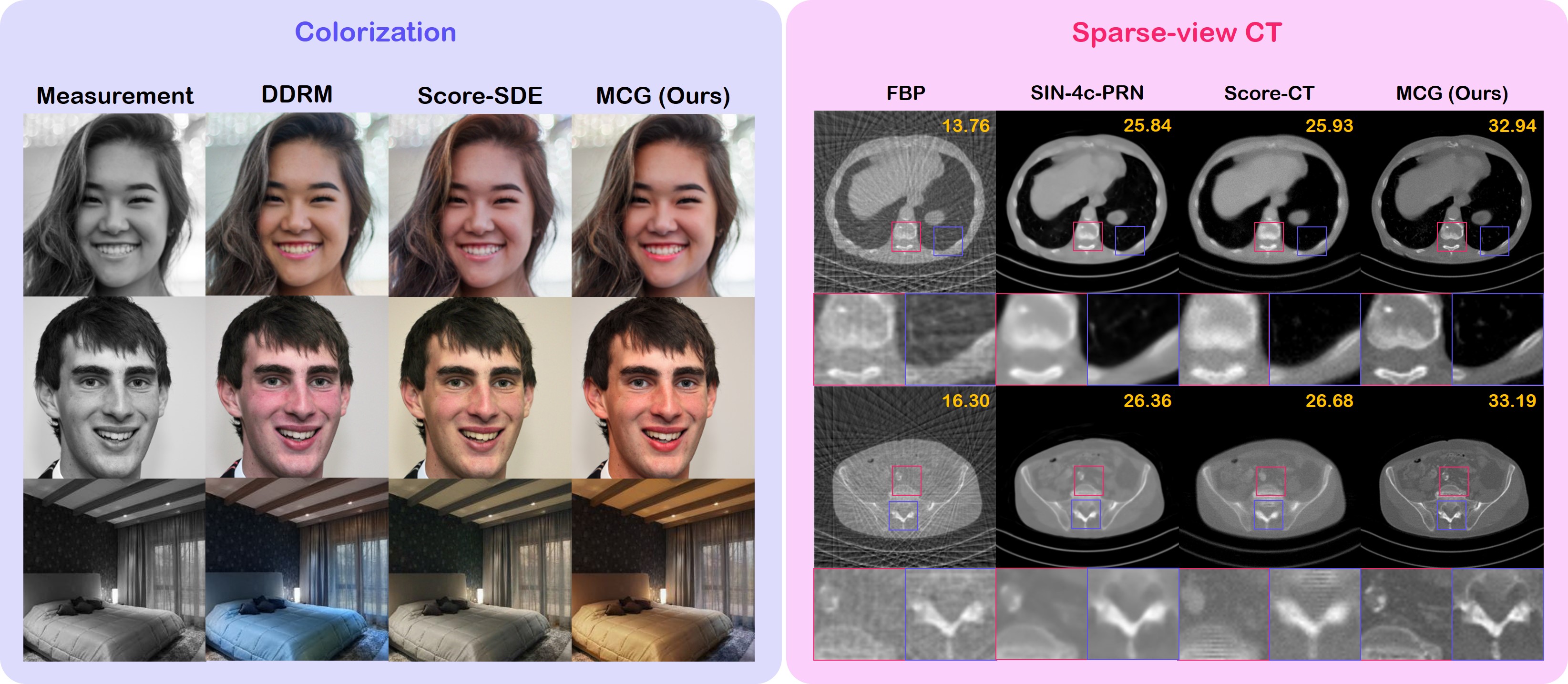}
    \caption{Colorization results on FFHQ / LSUN-bedroom, Sparse view CT reconstruction results on AAPM.}
    \label{fig:results_color_ct}
\end{figure}

\paragraph{Colorization}

We choose score-SDE~\cite{song2020score}, and DDRM~\cite{kawar2022denoising} as diffusion-model based comparison methods, and also compare against cINN~\cite{ardizzone2019guided}, and pix2pix~\cite{isola2017image}. Two metrics were used for evaluation: structural similarity index (SSIM), and LPIPS. Consistent with the findings from inpainting, we achieve much improved performance than score-SDE, and also is favorable against state-of-the-art (SOTA) superivsed learning based methods.
In Table~\ref{tab:comparison-color}, we see that the proposed method outperforms all other methods in terms of both PSNR/LPIPS in LSUN-bedroom, and also achieves strong performance in the colorization of FFHQ dataset.

\paragraph{CT reconstruction}

To the best of our knowledge, \cite{song2022solving} is the only method that tackles CT reconstruction directly with diffusion models. We compare our method against \cite{song2022solving}, which we refer to as score-CT henceforth. We also compare with the best-in-class supervised learning methods, cGAN~\cite{ghani2018deep} and SIN-4c-PRN~\cite{wei20202}. As a compressed sensing baseline, FISTA-TV~\cite{beck2009fast} was included, along with the analytical reconstruction method, FBP. We use two standard metrics - peak-signal-to-noise-ratio (PSNR), and SSIM for quantitative evaluation.
From Table~\ref{tab:comparison-ct}, we see that the newly proposed MCG method outperforms the previous score-CT~\cite{song2022solving} by a large margin. We can observe the superiority of MCG over other methods more clearly in Fig.~\ref{fig:results_color_ct}, where MCG reconstructs the measurement with high fidelity and detail. All other methods including the fully supervised baselines fall behind the proposed method.

\begin{wraptable}[]{r}{0.4\textwidth}
\centering
\setlength{\tabcolsep}{0.2em}
\resizebox{0.30\textwidth}{!}{
\begin{tabular}{lc}
\toprule
Method & Wall-clock time [s]\\
\midrule
{Score-SDE~\cite{song2020score}} & {38.68} \\
{RePAINT~\cite{lugmayr2022repaint}} & {247.6} \\
{DDRM~\cite{kawar2022denoising}} & {2.117} \\
{LaMa~\cite{suvorov2022resolution}} & {0.629} \\
{AOT-GAN~\cite{zeng2022aggregated}} & {0.082} \\
{ICT~\cite{wan2021high}} & {144.6} \\
{DSI~\cite{peng2021generating}} & {36.64} \\
{IAGAN~\cite{hussein2020image}} & {518.47} \\
\midrule
{Ours} & {81.59} \\
\bottomrule
\end{tabular}
}
\vspace{0.2em}
\caption{
Runtime for each algorithm in Wall-clock time: Computed with a single GTX 1080Ti GPU.
}
\label{tab:runtime}
\end{wraptable}

\paragraph{Ablation studies}

We perform three ablation studies: 1) As both the MCG term and the projection term contain information about the measurement $\y$, we observe the contribution of each term to the fixed solution. To further clarify the efficacy of the gradient step combined with Tweedie's denoising, we also consider the case where the gradient of the log likelihood is computed not in the noiseless regime, but in the noise level matching the current iteration. Specifically, we define $\x'_{i-1} := \f(\x_i, \s_\theta) + g(\x_i)\z,\, \z \sim \Nc(0, \Ib)$ , $\y_{i-1} \sim p(\y_{i-1}|\y_0)$, and implement the gradient step as $\nabla_{\x_i}\|\y_{i-1} - \Hb\x'_{i-1}\|_2^2$. 2) As the performance of diffusion models depend heavily on the number of NFEs, we observe the trade-off of each diffusion model when varying the NFE from 20 to 1000. Moreover, for completeness, we measure the runtime of each algorithms including the non-diffusion based methods in wall-clock time computed with a commodity GPU in Table.~\ref{tab:runtime}.
3) Setting $\alpha = 0.0$ reduces our method to~\cite{chung2022come}. We show the difference in the performance by varying the values of $\alpha$.

\begin{wraptable}[6]{r}{0.5\textwidth}
\centering
\vspace{-0.7cm}
\setlength{\tabcolsep}{0.2em}
\resizebox{0.40\textwidth}{!}{
\begin{tabular}{lcc}
\toprule
Method & LPIPS($\downarrow$)       & MSE(MC)       \\
\midrule
Proj. & 0.138     & 0     \\
$\nabla_{\x_i}\|\y_{i-1} - \Hb\x'_{i-1}\|_2^2$ & 0.271 & 12.99 \\
$\nabla_{\x_i}\|\y_{i-1} - \Hb\x'_{i-1}\|_2^2$ + Proj. & 0.128 & 0 \\
$\nabla_{\x_i}\|\y - \Hb\hat\x_0\|_2^2$ & 0.124 & 10.7 \\
$\nabla_{\x_i}\|\y - \Hb\hat\x_0\|_2^2$ + Proj. (\textbf{Ours}) & \textbf{0.089} & \textbf{0} \\
\bottomrule
\end{tabular}
}
\vspace{0.2em}
\caption{LPIPS \& Measurement consistency (MC) vs. method}
\label{fig:ablation_method}
\end{wraptable}

First, we see in Table.~\ref{fig:ablation_method} that using only the MCG step leads to improved performance in terms of LPIPS, but introduces error in the measurement consistency (measured with MSE). Combining both the projection and MCG leads to perfect data consistency along with further improved reconstruction. When considering gradient steps without Tweedie's denoising (i.e. keeping the noise level at the $i^{\text{th}}$ step), the performance heavily degrades, especially when implemented without the projection steps. Here, we see that the proposed denoising step to utilize $\hat\x_0$ is indeed the key to the superior performance.

\begin{figure}
     \centering
     \begin{subfigure}[t]{0.45\textwidth}
         \centering
         \includegraphics[width=\textwidth]{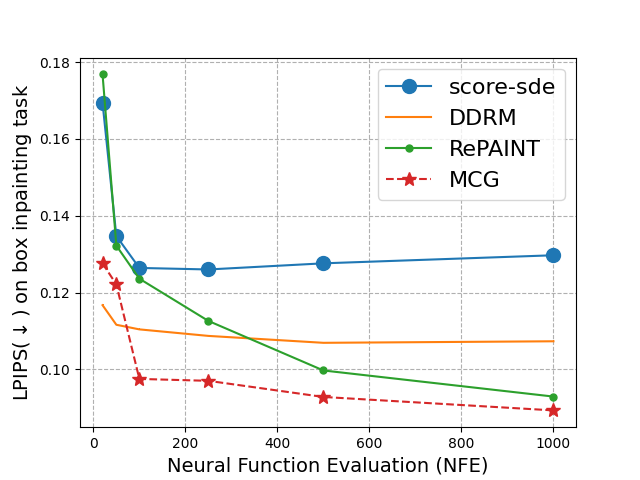}
         \caption{LPIPS vs. NFE}
         \label{fig:ablation_nfe}
     \end{subfigure}
     \hfill
     \begin{subfigure}[t]{0.45\textwidth}
         \centering
         \includegraphics[width=\textwidth]{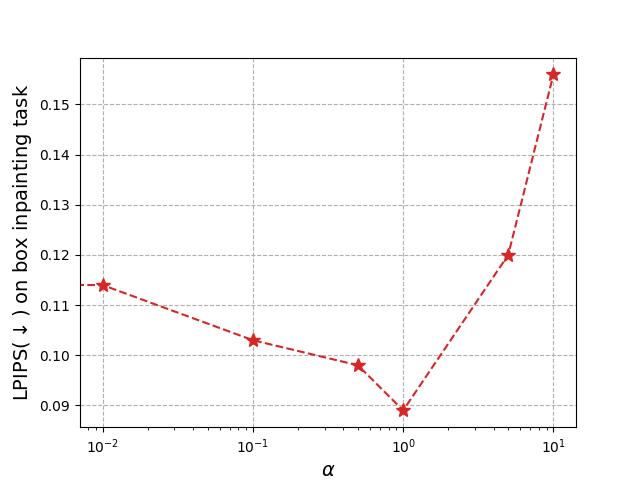}
         \caption{LPIPS vs. $\alpha'$}
         \label{fig:ablation_alpha}
     \end{subfigure}
        \caption{Ablation studies performed with box inpainting task on FFHQ 256$\times$256 data.}
        \label{fig:ablation}
\end{figure}

Second, looking at Fig.~\ref{fig:ablation_nfe}, we immediately see that the graph of MCG stays in the lowest (best) LPIPS regime across all NFEs by a large margin, except for when the NFE drops below 100. Here, DDRM~\cite{kawar2022denoising} takes over the 1st place - allegedly due to the DDIM sampling strategy they take. The performance of RePAINT deteriorates rapidly as we decrease NFE. Furthermore, we observe that the LPIPS of score-SDE~\cite{song2020score} actually {\em increases} (i.e. worsen), as we increase the number of NFEs from a few hundred to one thousand. This suggests that the inference process that score-SDE takes (i.e. projection only) is inherently flawed, and cannot be corrected by taking small enough steps. In Table.~\ref{tab:runtime}, we list the runtime of all the methods that were used for comparison in the task of inpainting. Note that the proposed method takes longer for compute than score-SDE albeit having the same NFE. The gap is due to the backpropagation steps that are required for the MCG step, where the gap can be potentially ameliorated by switching to JAX~\cite{jax2018github} implementation from the current PyTorch implementation.

Lastly, we observe the difference in the performance as we vary the values of $\alpha$. Implementation-wise, we find that we yield superior results when normalizing the squared norm with the norm of itself (e.g. $\alpha = \alpha' / \|\Wb(\y - \Hb\hat\x_0)\|$, where $\alpha'$ is some constant). In order to avoid cluttered notation, we instead experiment with changing the values of $\alpha'$ in Fig.~\ref{fig:ablation_alpha}.
Inspecting Fig.~\ref{fig:ablation_alpha}, we see that $\alpha$ values within the range $[0.1, 1.0]$ produce satisfactory results. $\alpha$ values that are too low do not fully enjoy the advantages of MCG and collapses to the projection-only method, while using too high values of $\alpha$ results in exploding gradients, and the reconstruction saturates.

\paragraph{Properties of our method}
Our proposed method is fully unsupervised and is not trained on solving a specific inverse problem. For example, our box masks and random masks have very different forms of erasing the pixel values. Nevertheless, our method generalizes perfectly well to such different measurement conditions, while other methods have a large performance gap between the different mask shapes. We further note two appealing properties of our method as an inverse problem solver: 1) the ability to generate multiple solutions given a condition, and 2) the ability to maintain perfect measurement consistency. The former ability often lacks in supervised learning-based methods~\cite{suvorov2022resolution,wei20202}, and the latter is often not satisfied for some unsupervised GAN-based solutions~\cite{daras2021intermediate,bora2017compressed}.

\section{Conclusion}
\label{sec:conclusion}

In this work, we proposed a general framework that can greatly enhance the performance of the diffusion model-based solvers for solving inverse problems. We showed several promising applications - inpainting, colorization, sparse view CT reconstruction, and showed that our method can outperform the current state-of-the-art methods. We analyzed our method theoretically and show that MCG prevents the data generation process from falling off the manifold, thereby reducing the errors that might accumulate at every step. Further, we showed that MCG controls the direction tangent to the data manifold, whereas the score function controls the direction that is normal, such that the two components complement each other.

\paragraph{Limitations and Broader Impact}

The proposed method is inherently stochastic since the diffusion model is the main workhorse of the algorithm. When the dimension $m$ is pushed to low values, at times, our method fails to produce high quality reconstructions, albeit being better than the other methods overall. For extreme cases of inpainting (e.g. Half masks) with the ImageNet model, we often observe artifacts in our reconstruction (e.g. generating perfectly symmetric images), which we discuss in further detail in Sec.~\ref{sec:limitation}. We note that our method is slow to sample from, inheriting the existing limitations of diffusion models. This would likely benefit from leveraging recent solvers aimed at accelerating the inference speed of diffusion models. In line with the arguments of other generative model-based inverse problem solvers, our method is a solver that relies heavily on the underlying diffusion model, and can thus potentially create malicious content such as deepfakes. Further, the reconstructions could intensify the social bias that is already existent in the training dataset.

\begin{ack}
This research was supported by Field-oriented Technology Development Project for Customs Administration through National Research Foundation of Korea(NRF) funded by the Ministry of Science \& ICT and Korea Customs Service (NRF-2021M3I1A1097938, NRF-2021M3I1A1097910), by the Korea Health Technology R\&D Project through the Korea Health Industry Development Institute (KHIDI), which is funded by the Ministry of Health \& Welfare, Republic of Korea (grant number: HU21C0222), and by the KAIST Key Research Institute (Interdisciplinary Research Group) Project.

\end{ack}

\bibliographystyle{plain}
\bibliography{reference}

\clearpage
\section*{Checklist}


\begin{enumerate}

\item For all authors...
\begin{enumerate}
  \item Do the main claims made in the abstract and introduction accurately reflect the paper's contributions and scope?
    \answerYes{}
  \item Did you describe the limitations of your work?
    \answerYes{We discuss the limitations in \eqref{sec:conclusion}.}
  \item Did you discuss any potential negative societal impacts of your work?
    \answerYes{We discuss potential negative impacts in \eqref{sec:conclusion}.}
  \item Have you read the ethics review guidelines and ensured that your paper conforms to them?
    \answerYes{}
\end{enumerate}

\item If you are including theoretical results...
\begin{enumerate}
  \item Did you state the full set of assumptions of all theoretical results?
    \answerYes{}
        \item Did you include complete proofs of all theoretical results?
    \answerYes{We provide all proofs of results in supplementary material.}
\end{enumerate}

\item If you ran experiments...
\begin{enumerate}
  \item Did you include the code, data, and instructions needed to reproduce the main experimental results (either in the supplemental material or as a URL)?
    \answerYes{We include all code for our experiments in the supplementary material. We will release the code once the paper is published.}
  \item Did you specify all the training details (e.g., data splits, hyperparameters, how they were chosen)?
    \answerYes{}
        \item Did you report error bars (e.g., with respect to the random seed after running experiments multiple times)?
    \answerNo{Due to our limited resources we do not have time to run multiple sets of experiments.}
        \item Did you include the total amount of compute and the type of resources used (e.g., type of GPUs, internal cluster, or cloud provider)?
    \answerYes{}
\end{enumerate}

\item If you are using existing assets (e.g., code, data, models) or curating/releasing new assets...
\begin{enumerate}
  \item If your work uses existing assets, did you cite the creators?
    \answerYes{We have cited the original works that released the datasets.}
  \item Did you mention the license of the assets?
    \answerNo{Licenses are standard and can be found online.}
  \item Did you include any new assets either in the supplemental material or as a URL?
    \answerYes{We include our implementation as the supplementary material. We will release the code upon publication.}
  \item Did you discuss whether and how consent was obtained from people whose data you're using/curating?
    \answerNo{All datasets used in our work are publicly available.}
  \item Did you discuss whether the data you are using/curating contains personally identifiable information or offensive content?
    \answerNA{}
\end{enumerate}

\item If you used crowdsourcing or conducted research with human subjects...
\begin{enumerate}
  \item Did you include the full text of instructions given to participants and screenshots, if applicable?
    \answerNA{}
  \item Did you describe any potential participant risks, with links to Institutional Review Board (IRB) approvals, if applicable?
    \answerNA{}
  \item Did you include the estimated hourly wage paid to participants and the total amount spent on participant compensation?
    \answerNA{}
\end{enumerate}

\end{enumerate}


\clearpage
\appendix

\section{Proofs}

First, we remind our notation and the assumption.

\textbf{Notation} For a scalar $a$, points $\x,\y$ and a set $A$, we use the following notations:
$aA:= \{a \x: \x \in A\}$;
$d(\x,A) := \inf_{\y \in A} ||\x-\y||_2$;
$B_r(A):= \{\x: d(\x, A) < r\}$; $T_\x \Mc$: the tangent space to a manifold $\Mc$ at $\x$; $\Jb_f$: the jacobian matrix of a vector valued function $f$.

\manifoldassumption*

We state our proofs below.

\noisymanifold*
\begin{proof}
    Suppose that the data manifold is an $l$-dimensional linear subspace. By rotation and translation, we safely assume that $\Mc = \{ \x \in \Rd^n: x_{l+1} = x_{l+2} = \dots = x_n = 0\}.$ Then, we can simply write $d(\x,\Mc) = \sqrt{x_{l+1}^2+\dots+x_{n}^2}$, and $\Mc_i = \{ \x \in \Rd^n : x_{l+1}^2+\dots+x_{n}^2 = r_i^2 \}$.
    For a given point $\x' = (x'_1, x'_2, \dots) \in \Mc$, we consider $p(\x|\x')\sim\Nc(a_i\x',b_i^2 I)$ and obtain a concentration inequality independent to the choice of $\x'$.
    We need the standard Laurent-Massart bound for a chi-square variable \cite{laurent2000adaptive}. When $X$ is a chi-square distribution with $k$ degrees of freedom,
    \begin{align*}
        P[X-k\ge 2\sqrt{kt}+2t]&\le e^{-t},\\
        P[X-k\le -2\sqrt{kt}]&\le e^{-t}.
    \end{align*}
    As $\frac{x_{l+1}^2}{b_i^2}+\dots+\frac{x_{n}^2}{b_i^2}$ is a chi-square distribution with $n-l$ degrees of freedom, by substituting $t = (n-l) \varepsilon'$ in the above bound,
    \begin{align*}
        P\left[-2 (n-l) \sqrt{\varepsilon'} \le \frac{x_{l+1}^2}{b_i^2}+\dots+\frac{x_{n}^2}{b_i^2} - (n-l) \le 2(n-l) (\sqrt{\varepsilon'}+ \varepsilon') \right]& \\
        = P\left[ \sqrt{ x_{l+1}^2+\dots+x_{n}^2 }\in (\,r_i \sqrt{1- 2\sqrt{\varepsilon'}},\, r_i \sqrt{1+ 2\sqrt{\varepsilon'}+2\varepsilon'}\,) \right] &\ge 1-2e^{-(n-l)\varepsilon'}.
    \end{align*}
    Note that the above inequality does not depend on $x_1, \dots x_l$, thus the choice of $\x' \in \Mc$.
    As a result, by setting $\varepsilon = \min\{ 1- \sqrt{1- 2\sqrt{\varepsilon'}}, \sqrt{1+ 2\sqrt{\varepsilon'}+2\varepsilon'} -1 \}$ and $\delta = 2e^{-(n-l)\varepsilon'}$,
    $$p(\x \in B_{\varepsilon r_i}(M_i) | \x') >1-\delta,$$ thus
    $$p_i(\x \in B_{\varepsilon r_i}(M_i) ) = \int p(\x \in B_{\varepsilon r_i}(M_i) | \x') p(\x') d\x' > 1-\delta.$$
\end{proof}
\score*
\begin{proof}
    To minimize \eqref{eq:dsm}, or equivalently,
    \begin{align*}
        \int ||s_\theta(\x_t,t) - \nabla_{\x_t} \log p(\x_t|\x_0)||_2^2 p(\x_t|\x)p(\x) d\x d\x_t dt,
    \end{align*}
    By differentiating the objective with respect to $s_\theta(\x_t,t)$, we have
    \begin{gather*}
        \int \left( s_\theta(\x_t,t) - \frac{a_t \x - \x_t}{b_t^2} \right) p(\x_t|\x)p(\x) d\x = 0 \\
        \int s_\theta(\x_t,t) p(\x_t) p(\x|\x_t) d\x = \int \frac{a_t \x - \x_t}{b_t^2} p(\x_t) p(\x|\x_t) d\x\\
        s_\theta(\x_t,t) \int p(\x|\x_t) d\x = \int \frac{a_t \x}{b_t^2} p(\x|\x_t) d\x - \frac{\x_t}{b_t^2} \int p(\x|\x_t) d\x \\
        \therefore s_\theta(\x_t,t) = \frac{1}{b_t^2} ( -\x_t + a_t  \int \x p(\x|\x_t) d\x ) \forall \x_t,t,
    \end{gather*}
    where we used $p(\x_t|\x)p(\x) = p(\x,\x_t) = p(\x_t)p(\x|\x_t)$, $p(\x_t)>0$, and $\int p(\x|\x_t) d\x = 1$ in each line.
    Here, $Q_i(\x_i) = \int \x p(\x|\x_i) d\x$ is the weighted average vector of points on the data manifold as $p(\x|\x_i)$ is supported on the data manifold.
    Combining it with the assumption that the manifold is linear, $Q_i(\x_i) \in \Mc$.
    
    Considering the symmetry of $p(\x|\x_i)$ about $\x_i$, $p(\x|\x_i)$ is a radial function on $\Mc$, centering around the nearest point to $\x_i$ on $\Mc$.
    Hence, $Q_i(\x_i)$ shall be the nearest point to $\x_i$ of all points on $\Mc$.
    Therefore, $J_{Q_i}$ is the orthogonal projection onto $T_{Q_i(\x_i)}\Mc$.
    Stating more rigorously, let $\ub = \ub_t + \ub_n \in \Rd^n$ for $\ub_t \in T_{Q_i(\x_i)}\Mc, \ub_n \perp T_{Q_i(\x_i)}\Mc$.
    Then, for a scalar $s$, $Q_i(\x_i + s\ub) = Q_i(\x_i) + s\ub_t$, as only tangent component to the manifold change the nearest point.
    By differentiating with respect to $s$, we obtain $\Jb_{Q_i} \ub = \ub_t$, thus $\Jb_{Q_i}^2 = \Jb_{Q_i}$.
    For another vector $\vb = \vb_t + \vb_n$ with $\vb_t \in T_{Q_i(\x_i)}\Mc, \vb_n \perp T_{Q_i(\x_i)}\Mc$,
    \begin{align*}
        \vb^T \Jb_{Q_i} \ub &= (\vb_t+\vb_n)^T \ub_t\\
        &= \vb_t^T \ub_t\\
        &= (\ub_t + \ub_n)^T \vb_t\\
        &= \ub^T \Jb_{Q_i} \vb,
    \end{align*}
    where we applied $\vb_n^T \ub_t = 0 = \ub_n^T \vb_t$.
    Therefore, $\Jb_{Q_i}$ is symmetric, i.e. $\Jb_{Q_i}^T = \Jb_{Q_i}$, which concludes this proof.
\end{proof}
\Improved*
\begin{figure*}[t!]
    \centering
    \includegraphics[width=0.45\textwidth]{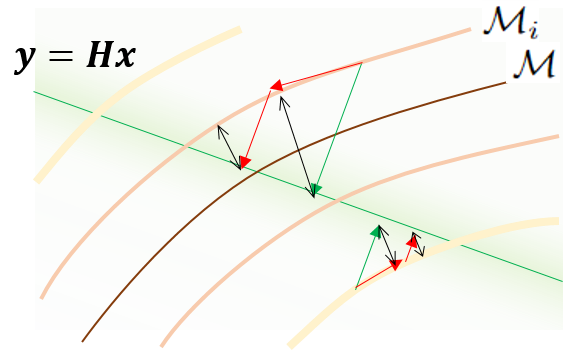}
	\caption{The advantage of mixing the MCG and the POCS steps over the conventional POCS step.
	Each curve represents a manifold of  (noisy) data.
	Arrows suggest the POCS steps (green arrows) and steps mixing the MCG and the POCS (red arrows). Due to the path along the manifolds, proposed mixing step alleviates reverse diffusion step leaving the manifolds (black arrows).
	}
	\label{fig:comparison}
\end{figure*}
\begin{proof}
\begin{align*}
    \frac{\partial}{\partial\x_i} \|\Wb (\y - \Hb\hat\x_0)\|_2^2 &= - 2 \Jb_{\Wb \Hb Q_i}^T \Wb (\y - \Hb\hat\x_0)\\
    &= -2 \Jb_{Q_i}^T \Hb^T \Wb^T \Wb (\y - \Hb\hat\x_0) \\
    & = \Jb_{Q_i} d \in T_{Q_i(\x_i)}\Mc
\end{align*}
where $d = -2\Hb^T \Wb^T \Wb (\y - \Hb\hat\x_0)$.
The first and second equality is given by the chain rule and the last line is by Proposition~\ref{prop:score}.
\end{proof}
In Fig~\ref{fig:comparison}, we illustrate how the proposed algorithm benefits from mixing the MCG step with the conventional POCS step.
Pushing the points to the tangent directions, we expect less deviation from the manifold which is attributed to POCS.

\section{Discrete forms of SDE}
\label{sec:app-sde}

Here, we review the different types of SDEs and sampling algorithms that we use throughout the paper for completeness. We assume that the time horizon $[0, 1]$ is linearly split up into $N$ discretization segments, such that all intervals have the length $1/N$, if not specified otherwise.

\subsection{Forward diffusion}
Due to the linearity of the drift and diffusion functions, we can analytically sample from $p(\x_i|\x_0)$ via reparameterization trick:
\begin{align}
    \x_i = a_i \x_0 + b_i \z,\, \z \sim \Nc(0, \Ib).
\end{align}
In VP-SDE~\cite{ho2020denoising}, one defines a linearly increasing noise schedule $\beta_1, \beta_2, \dots, \beta_N \in (0, 1)$. Further, we define $\alpha_i = 1 - \beta_i$, and $\bar\alpha_i = \prod_{j=1}^{i}\alpha_j$. Then, the forward diffusion process can be implemented as
\begin{align}
\label{eq:vp-f}
    \x_i = \sqrt{\bar\alpha_i}\x_0 + \sqrt{1 - \bar\alpha_i}\z,\,\z \sim \Nc(0, \Ib).
\end{align}
In VE-SDE~\cite{song2020score}, one defines a geometrically increasing noise schedule $\sigma_i = \sigma_0\left(\frac{\sigma_N}{\sigma_0}\right)^{\frac{i-1}{N-1}}$. Since the drift function is zero, the forward diffusion simply becomes Brownian motion. Concretely,
\begin{align}
\label{eq:ve-f}
    \x_i = \x_0 + \sigma_i \z,\, \z \sim \Nc(0, \Ib).
\end{align}

\subsection{Reverse diffusion}
First, for the case of VP-SDE, the reverse diffusion step is implemented by
\begin{align}
\label{eq:vp-r-z}
    \x_{i-1} = \frac{1}{\sqrt{\alpha_i}}\left(\x_i - \frac{1 - \alpha_i}{\sqrt{1 - \bar\alpha_i}} \z_\theta(\x_i,i)\right) + \sqrt{\tilde\sigma_i}\z,\,~~\z \sim \Nc(0, \Ib),
\end{align}
where $\z_\theta(\x_i,i)$ is trained with the epsilon-matching scheme as in~\cite{ho2020denoising}, and $\tilde\sigma_i$ is set to a learnable parameter as in~\cite{dhariwal2021diffusion}. Note that eq.~\eqref{eq:vp-r-z} was written in terms of $\z_\theta(\x_i,i)$ and not in terms of the score function, $\s_\theta(\x_i,i)$. One can re-write the expression using the relation $\z_\theta(\x_i,i) = -\sqrt{1 - \bar\alpha_i}\s_\theta(\x_i,i)$, as
\begin{align}
\label{eq:vp-r-s}
    \x_{i-1} = \frac{1}{\sqrt{\alpha_i}}\left(\x_i + (1 - \alpha_i) \s_\theta(\x_i,i)\right) + \sqrt{\sigma_i}\z,\,\z \sim \Nc(0, \Ib).
\end{align}
Next, for the VE-SDE, the reverse diffusion step using the Euler-Maruyama solver~\cite{sarkka2019applied} is given as
\begin{align}
    \x_{i-1} = \x_i + (\sigma_i^2 - \sigma_{i-1}^2)s_\theta(\x_i,i) + \sqrt{\sigma_i^2 - \sigma_{i-1}^2}\z,\,\z \sim \Nc(0, \Ib).
\end{align}
Summary is presented in Table~\ref{tab:app-abfg}.

\begin{table*}[]
\centering
\setlength{\tabcolsep}{0.2em}
\renewcommand{\arraystretch}{1.5}
\resizebox{0.7\textwidth}{!}{
\begin{tabular}{c|cccc}
\toprule
Type & $a_i$ & $b_i$ & $\f(\x_i, s_\theta)$ & $g(i)$ \\
\hline
VP-SDE & $\sqrt{\bar\alpha_i}$ & $\sqrt{1 - \bar\alpha_i}$ &  $\frac{1}{\sqrt{\bar\alpha_i}}\left(\x_i + (1 - \alpha_i)s_\theta(\x_i,i)\right)$
 &  $\sqrt{\tilde\sigma_i}$ \\
VE-SDE & 1 & $\sigma_i$ & $\x_i + (\sigma_i^2 - \sigma_{i-1}^2)s_\theta(\x_i, i)$ & $\sqrt{\sigma_i^2 - \sigma_{i-1}^2}$ \\
\bottomrule
\end{tabular}
}
\vspace{0.2em}
\caption{
Choice of $a_i, b_i, \f, g$ for each SDE realization.
}
\label{tab:app-abfg}
\end{table*}

\section{Algorithms}
\label{sec:app-algo}

\begin{algorithm}[!hbt]
\caption{Inpainting (VP, AS)}
\begin{algorithmic}[1]
\Require {${\y}, \Pb, \{\alpha_i\}_{i=1}^{N}, \{\tilde\sigma_i\}_{i=1}^{N}, s_{\theta}$, $\alpha$}
\State $\x_N \sim \Nc(\textbf{0}, \Ib)$ \Comment{Initial sampling}
\For{$i = N$ to $1$} \Comment{Reverse diffusion} \do \\
\State $s \gets s_{\theta}(\x_i,i)$ \Comment{Cache score function output}
\State $\x'_{i-1} \gets \frac{1}{\sqrt{\alpha_i}}(\x_i +{(1 - \alpha_i)}s)$
\State $\z \sim \Nc(\textbf{0}, \Ib)$
\State $\x_{i-1} \gets \x'_{i-1} + \tilde\sigma_i\z$ \Comment{Unconditional update}
\State $\z \sim \Nc(\textbf{0}, \Ib)$
\State $\hat{\x}_0 \gets \frac{1}{\sqrt{\bar\alpha_i}}(\x_i + (1 - \bar\alpha_i) s)$ \Comment{$\hat{\x}_0$ prediction}
\State $\y_i \gets \sqrt{\bar{\alpha}_i}\y + \sqrt{1 - \bar{\alpha}_i}\z$
\State $\x''_{i-1} \gets \x'_{i-1} - \alpha\frac{\partial}{\partial \x_i}\|\y - \Pb \hat\x_0\|_2^2$
    \Comment{MCG}
\State $\z \sim \Nc(\textbf{0}, \Ib)$
\State $\y_i \gets \sqrt{\bar\alpha_i}\y + \sqrt{1 -\bar\alpha_i}\z$
\State $\x_{i-1} \gets (\Ib - \Pb^T\Pb)\x''_{i-1} + \Pb^T\y_i$
\Comment{Data consistency}
\EndFor
\State \textbf{return} {$\x_0$}
\end{algorithmic}\label{alg:MCG_inpaint}
\end{algorithm}

\paragraph{Inpainting}

The forward model for inpainting is given as
\begin{align}
\label{eq:fm-inpaint}
    \y = \Pb\x + \bm{\epsilon},\quad \Pb \in \Rd^{m \times n},
\end{align}
where $\Pb \in \{0,1\}^{m \times n}$ is the matrix consisting of the columns with standard coordinate vectors indicating the indices of measurement.
For the steps in \eqref{eq:reverse_discrete_ip_mcg}, \eqref{eq:nem_mcg},  we choose the following
\begin{align}
\label{eq:choice-inpaint}
    \Wb = \Ib,\quad \Ab = \Ib - \Pb^T\Pb,\quad \bb_i = \Pb^T\y_i,\,\quad \y_i \sim q(\y_i|\y) := \Nc(\y_i|a_i\y, b_i^2 \Ib).
\end{align}
Specifically, $\Ab$ takes the orthogonal complement of $\x'_{i-1}$, meaning that the measurement subspace is corrected by $\y_i$, while the orthogonal components are updated from $\x'_{i-1}$. Note that we use $\y_i$ sampled from $\y$ to match the noise level of the current estimate.

We provide the algorithm used for inpainting in Algorithm.~\ref{alg:MCG_inpaint}. The sampler is based on basic ancestral sampling (AS) of~\cite{ho2020denoising}, and the default configuration requires $N = 1000$, $\alpha=1.0 / \|\y - \Pb\hat\x_0\|$ for sampling.

\paragraph{Colorization}

The forward model for colorization is specified as
\begin{equation}
\label{eq:fm-color}
    \y = \Cb\x + \bm{\epsilon} := \Pb\Mb\x + \bm{\epsilon},\quad \Pb \in \Rd^{m \times n},\quad \Mb \in \Rd^{n \times n},
\end{equation}
where $\Pb$ is the matrix that was used in inpainting, and $\Mb$ is an orthogonal matrix that couples the RGB colormaps\footnote{The matrix $\Mb$ is adopted from the colorization matrix of~\cite{song2020score}.}. $\Mb^T$ is a matrix that de-couples the channels back to the original space. In other words, one can view colorization as performing imputation in some spectral space. Subsequently, for our colorization method we choose
\begin{equation}
\label{eq:choice-color}
    \Wb = \Cb^T,\quad \Ab = \Ib - \Cb^T\Cb,\quad \bb_i = \Cb^T\y_i,\ \quad \y_i \sim q(\y_i|\y).
\end{equation}
Again, our forward measurement matrix is orthogonal, and we choose $\Ab$ such that we only affect the orthogonal complement of the measurement subspace.

The sampler for colorization is based on the predictor-corrector (PC) sampler of~\cite{song2020score} (VE-SDE), and we choose to apply MCG after every iteration of both predictor, and corrector steps. $N = 2000, \alpha = 0.1 / \|\Cb^T(\y - \Cb\hat\x_0)\|$ are chosen as hyper-parameters.

\paragraph{CT Reconstruction}

For the case of CT reconstruction, the forward model reads
\begin{equation}
\label{eq:fm-ct}
    \y = \Rb\x + \bm{\epsilon},\quad \Rb \in \Rd^{m \times n},
\end{equation}
where $\Rb$ is the discretized Radon transfrom~\cite{buzug2011computed} that measures the projection images from different angles. Note that for CT applications, $\Rb^T$ corresponds to performing backprojection (BP), and $\Rb^{\dagger}$ corresponds to performing filtered backprojection (FBP). We choose
\begin{equation}
\label{eq:choice-ct}
    \Wb = \Rb^\dagger,\quad \Ab = \Ib - \Rb^T(\Rb\Rb^T)^\dagger \Rb,\quad \bb_i = \Rb^T(\Rb\Rb^T)^\dagger\y_i,\,\quad  \y_i \sim q(\y_i|\y),
\end{equation}
where the choice of $\Ab$ reflects that the Radon transform is not orthogonal, and we need the term $(\Rb\Rb^T)^\dagger$ as a term analogous to the filtering step. Indeed, this form of update is known as the algebraic reconstruction technique (ART), a classic technique in the context of CT~\cite{gordon1970algebraic}. We note that this choice is different from what was proposed in \cite{song2022solving}, where the authors repeatedly apply projection/FBP by explicitly replacing the sinogram in the measured locations. From our experiments, we find that repeated application of FBP is highly numerically unstable, often leading to overflow. This is especially the case when we have limited resources for training data (we use ~4k, whereas \cite{song2022solving} uses ~50k), as we further show in Section~\ref{sec:exp}.

Algorithm for SV-CT reconstruction uses PC sampler (VE-SDE), where we use MCG step after one sweep of corrector-predictor update. We note that this is a design choice, and one may as well use the MCG update step after both the predictor and corrector steps, as was proposed in~\cite{song2020score}. We set $N = 2000, \alpha = 0.1 / \|\Rb^\dagger(\y - \Rb \hat\x_0)\|$.

\begin{figure}[t!]
    \centering
    \includegraphics[width=\textwidth]{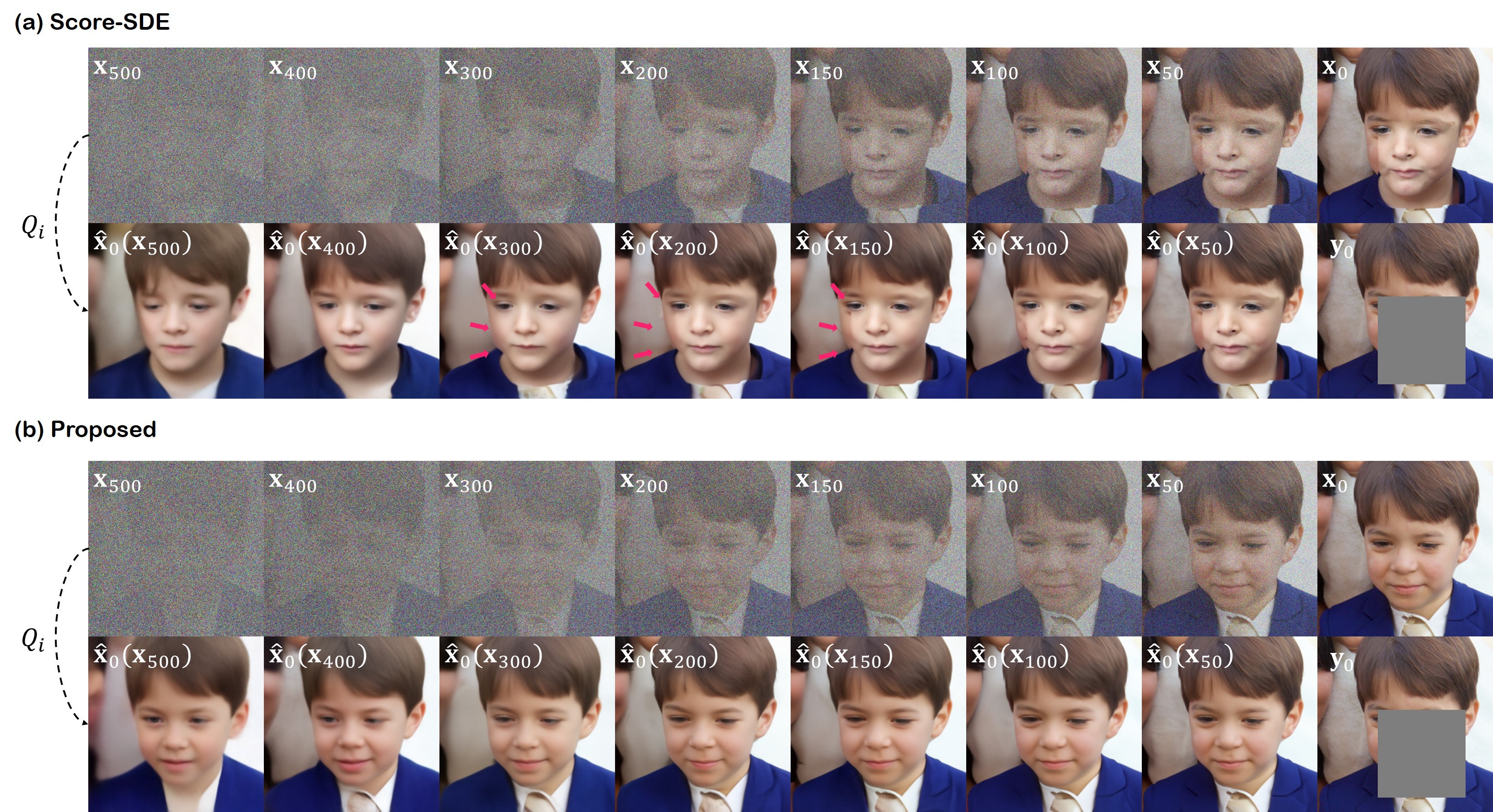}
	\caption{Comparison of the evolution (i.e. generative path) between score-SDE~\cite{song2020score}, and our method. First rows in (a),(b): Evolution of $\x_i$, second rows in (a),(b): Evolution of $\hat\x_0$.
	}
	\label{fig:evolution}
\end{figure}

\section{Generative process of the proposed method}
\label{sec:evolution}

In Fig.~\ref{fig:evolution}, we depict the comparison of the generative process between the two methods: score-SDE~\cite{song2020score}, which relies on alternating projections; and our method, which utilizes MCG as correcting steps. In Fig.~\ref{fig:evolution} (a), we can clearly see the unnatural boundary between the masked and the unmasked region forming, and evolving as $t \rightarrow 0$, without getting corrected (Visible more clearly in $\hat\x_0$). On the other hand, thanks to the additional gradient step that {\em corrects} the errors in the boundary, we see a much more natural evolution of the signal as $t \rightarrow 0$ in Fig.~\ref{fig:evolution} (b).

\begin{wrapfigure}[19]{r}{0.5\textwidth}
  \begin{center}
    \includegraphics[width=0.48\textwidth]{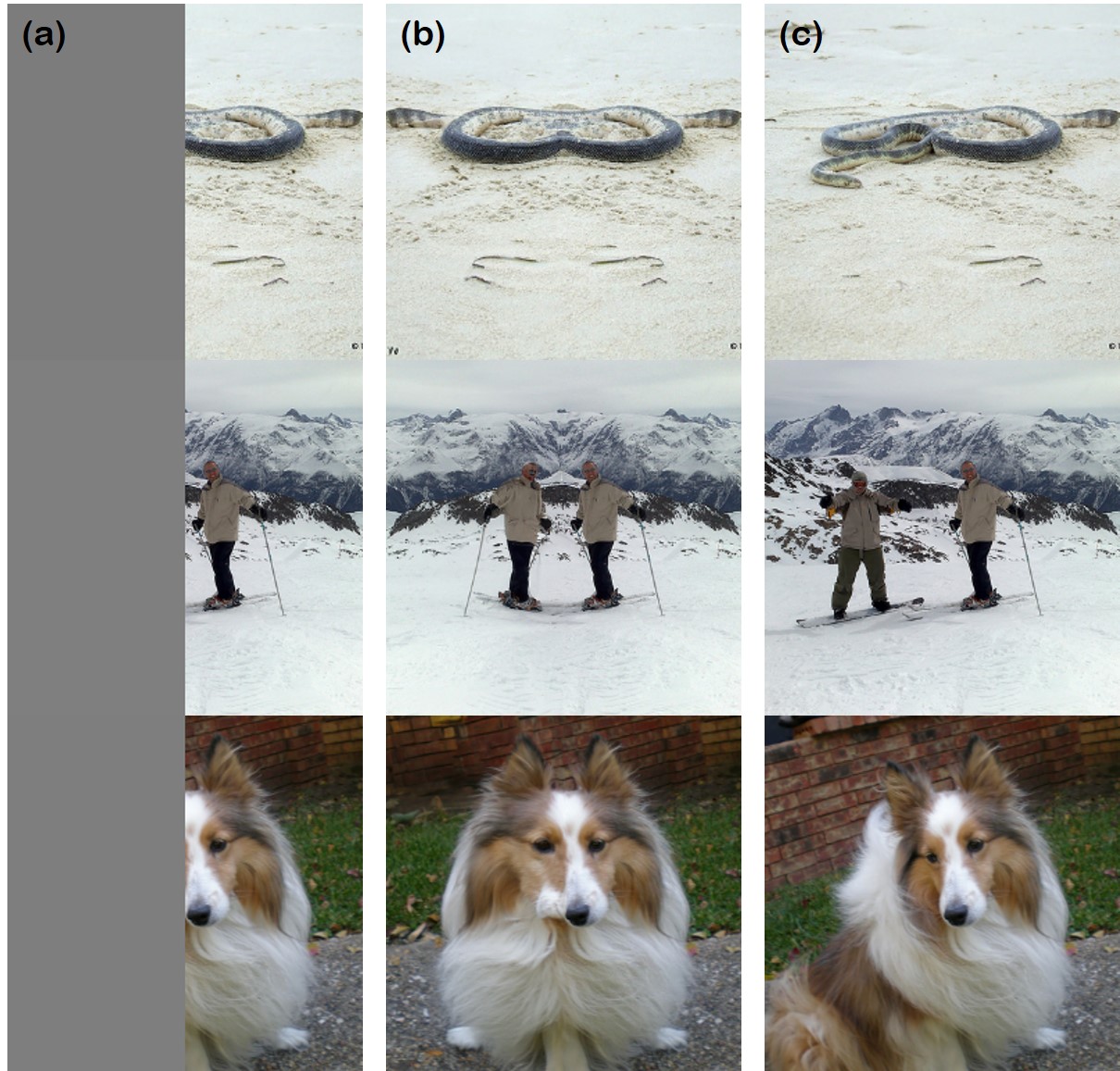}
  \end{center}
  \caption{Limitations of the proposed method. (a) Measurement, (b) reconstruction with the proposed method, (c) ground truth.}
  \label{fig:limitation}
\end{wrapfigure}

\section{Limitations}
\label{sec:limitation}

There exists a limitation specifically for the ImageNet dataset when using the proposed algorithm for inpainting. Specifically, as shown
in Fig.~\ref{fig:limitation},
for the case of half-mask (i.e. the left or right half of the image is zeroed-out), we often see the reconstructions are generated showing symmetries that are unrealistic. Note that this kind of effect is not observed in our FFHQ experiments. Hence, we conjecture that this phenomenon arises from the imperfectness of the learned score function $s_\theta$. Namely, due to the ImageNet dataset being much more diverse and therefore widely known to be a much harder dataset to learn, the subobtimality of the score function may be greater than the FFHQ score function. This could possibly lead to such deficiencies.

\section{Experimental Details}
\label{sec:app_exp_detail}

\subsection{Implementation details}

\paragraph{Training of the score function}

For inpainting experiments, we take the pre-trained score functions that are available online (FFHQ\footnote{\url{https://github.com/jychoi118/ilvr_adm}}, imagenet\footnote{\url{https://github.com/openai/guided-diffusion}}). For CT reconstruction experiment, we train a \texttt{ncsnpp} model with default configurations as guided in~\cite{song2020score} with the VE-SDE framework. The model was trained for 200 epochs with the full training dataset, with a single RTX 3090 GPU. Training took about one week wall-clock time.

\paragraph{Required compute time for inference}

All our sampling steps detailed in Algorithm~\ref{sec:app-algo} was performed with a single RTX 3090 GPU. The inpainting algorithm based on ADM~\cite{dhariwal2021diffusion} takes about 90 seconds (1000 NFE) to reconstruct a single image of size 256$\times$256. Our colorization and CT reconstruction algorithm based on score-SDE~\cite{song2020score} takes about 600 seconds (4000 NFE) to infer a single 256$\times$256 image.

\paragraph{Code Availability}

We will open-source our code used in our experiments upon publication to boost reproducibility.

\subsection{Comparison methods}

\subsubsection{Inpainting, Colorization}

\paragraph{Score-SDE}

Score-SDE~\cite{song2020score} demonstrated that unconditional diffusion models can be adopted to various inverse problems, such as inpainting and colorization. Our method without the MCG step is identical to score-SDE, and hence we use the same score function, parameters, and sampler as used in the proposed method for reconstruction.

\paragraph{RePAINT}

RePAINT~\cite{lugmayr2022repaint} proposes to iterate between denoising-noising steps multiple times in order to better incorporate inter-dependency between the known and the unknown regions in the case of image inpainting.
We use the same score function and sampler for RePAINT as in the proposed method. Following the default configurations in~\cite{lugmayr2022repaint}, we take $N = 200$ (corresponding to $T$ in~\cite{lugmayr2022repaint}), and $U = 10$, where $U$ denotes the count of iterated denoising-noising steps used within a single update index $i$.

\paragraph{DDRM}

DDRM~\cite{kawar2022denoising} demonstrates that linear inverse problems can be solved via diffusion models by decomposing the generative process with singular value decomposition (SVD), and performing reverse diffusion sampling in the spectral space.
The same score function adopted for the proposed method is used. Using the notations from~\cite{kawar2022denoising}, we choose $\sigma_{\y} = 0$, as we are aiming to solve noiseless inverse problem, and $\eta = 0.85, \eta_b = 1$. The number of NFE is set to 20 with the DDRM sampling steps.

\paragraph{LaMa}
LaMA contains fast Fourier convolution in generator architecture for reconstructing images. We trained the model from scratch using adversarial loss with r1 regularization term with its coefficient 10 and gradient penalty coefficient 0.001. Adam optimizer is used with the fixed learning rate of 0.001 and 0.0001 for discriminator network. For FFHQ and Imagenet dataset, 500k iterations of trainings were done with batch size of 8.

\paragraph{AOT-GAN}
AOT-GAN consists of a deep image generator with a AOT block which consists of multiple length of residual blocks in parallel. The discriminator is the same architecture with PatchGAN from ~\cite{zhu2017unpaired}. We trained the model from the scratch with 0.0001 learning rate using Adam optimizer $\beta_1 = 0$ and $\beta_2 = 0.9$ for both FFHQ and Imagent dataset. 500k iterations of trainings were done with batch size of 8. Also, for style loss and the perceptual loss, VGG19~\cite{simonyan2014very} pretrained on ImageNet~\cite{deng2009imagenet} was used.

\paragraph{ICT}
Image completion transformer (ICT) consists of two modules - a transformer model that follows the tokenization procedure to process information in the lower dimensional space, and another guided upsampling module to retrieve the data dimensionality. The encoded features are sampled from a probability distribution via Gibbs sampling, such that one can capture multimodal reconstructions from the same measurement.  
 For both the FFHQ and Imagenet dataset, we used pretrained models provided by the authors.

\paragraph{IAGAN}
Image adaptive GAN (IAGAN) uses a pre-trained generator and adapts it at test time for the given forward model. Specifically, following compressed sensing using generative model (CSGM)~\cite{bora2017compressed}, one initializes the latent vector $\z$ such that $\z^* = \argmin_\z \|y - AG_\theta(\z)\|$. Then, the latent code and the neural network parameters are jointly optimized through some iterations of $\z^{**},\theta^* = \argmin_{\z,\theta} \|y - AG_\theta(\z)\|$. The final result is achieved by the forward pass through the generator, after which follows the projection into the measurement subspace. For tuning the generator, we follow the default configurations from the official codebase. Since the codebase uses a GAN that generates 1024$\times$1024 images, we downscale the result into 256$\times$256 image as a final post-processing step.

\paragraph{DSI}
DSI is structured with the combination of VQ-VAE~\cite{van2017neural}, structure generator and texture generator. The architectures were trained separately, with Adam optimizer. When inference, only structure and texture generator was used. We trained the model from scratch. During optimization, the structure generator used linear warm-up schedular and square-root decay schedule used in ~\cite{razavi2019generating}. We used Adam optimizer on training all models with learning rate of 0.0001 and $\beta_1 = 0.5$ using exponetial moving average (EMA). Training was done for 500k iteration for both FFHQ and Imagenet dataset.

\paragraph{cINN}
cINN is an invertible neural network which can take in additional conditions as input, and in our case grayscale images. We train the model using default configurations as advised in \url{https://github.com/VLL-HD/conditional_INNs} without modifications. FFHQ model was trained with the learning rate of 0.0001 for 100 epohcs using the Adam optimizer. LSUN bedroom model was trained with the learning rate of 0.0001 for 30 epochs.

\paragraph{pix2pix}
Pix2pix is a variant of conditional GAN (cGAN) that takes in as input, the corrupted image. The model is trained in a supervised fashion, with the loss consisting of the reconstruction loss, and the adversarial loss. As the discriminator architecture, we adopt patchGAN~\cite{isola2017image}, and utilize the LSGAN~\cite{mao2017least} loss, weighting the adversarial loss by the value of 0.1. Similar to cINN, FFHQ model was trained with the learning rate of 0.0001 for 100 epochs using Adam optimizer. LSUN bedroom model was trained with the same configuration for 30 epochs.

\subsection{CT reconstruction}

\paragraph{Score-CT}

We use the hyper-parameters as advised in~\cite{song2022solving} and set $\eta = 0.246, \lambda = 0.841$. The measurement consistency step is imposed after every corrector-predictor sweep as in the proposed method.

\paragraph{SIN-4c-PRN}

Directly using the official implementation\footnote{\label{github}\url{https://github.com/anonyr7/Sinogram-Inpainting}}~\cite{wei20202}, we train the sinogram inpainting network (SIN) with the AAPM dataset for 200 epochs with the batch size of 8, and learning rate of 0.0001. We train two models separately for different number of views - 18, and 30.

\paragraph{cGAN}

We adopt the implementation of cGAN~\cite{ghani2018deep} from SIN-4c-PRN repository \footnoteref{github}. We train the two separate networks for 18 view, and 30 view projection, with the same configuration - 200 epochs, learning rate of 0.0001, and batch size of 8.

\paragraph{FISTA-TV}

We perform FISTA-TV~\cite{beck2009fast} reconstruction using \texttt{TomoBAR}~\cite{tooldkazanc}, together with the \texttt{CCPi} regularization toolkit~\cite{kazantsev2019ccpi}. Leveraging the default setting, we use the least-squares (LS) data model, and run the FISTA iteration for 300 iterations per image, with the total variation regularization strength set to 0.001.

\section{Further Experimental Results}

\begin{figure}[t]
    \centering
    \includegraphics[width=1.0\textwidth]{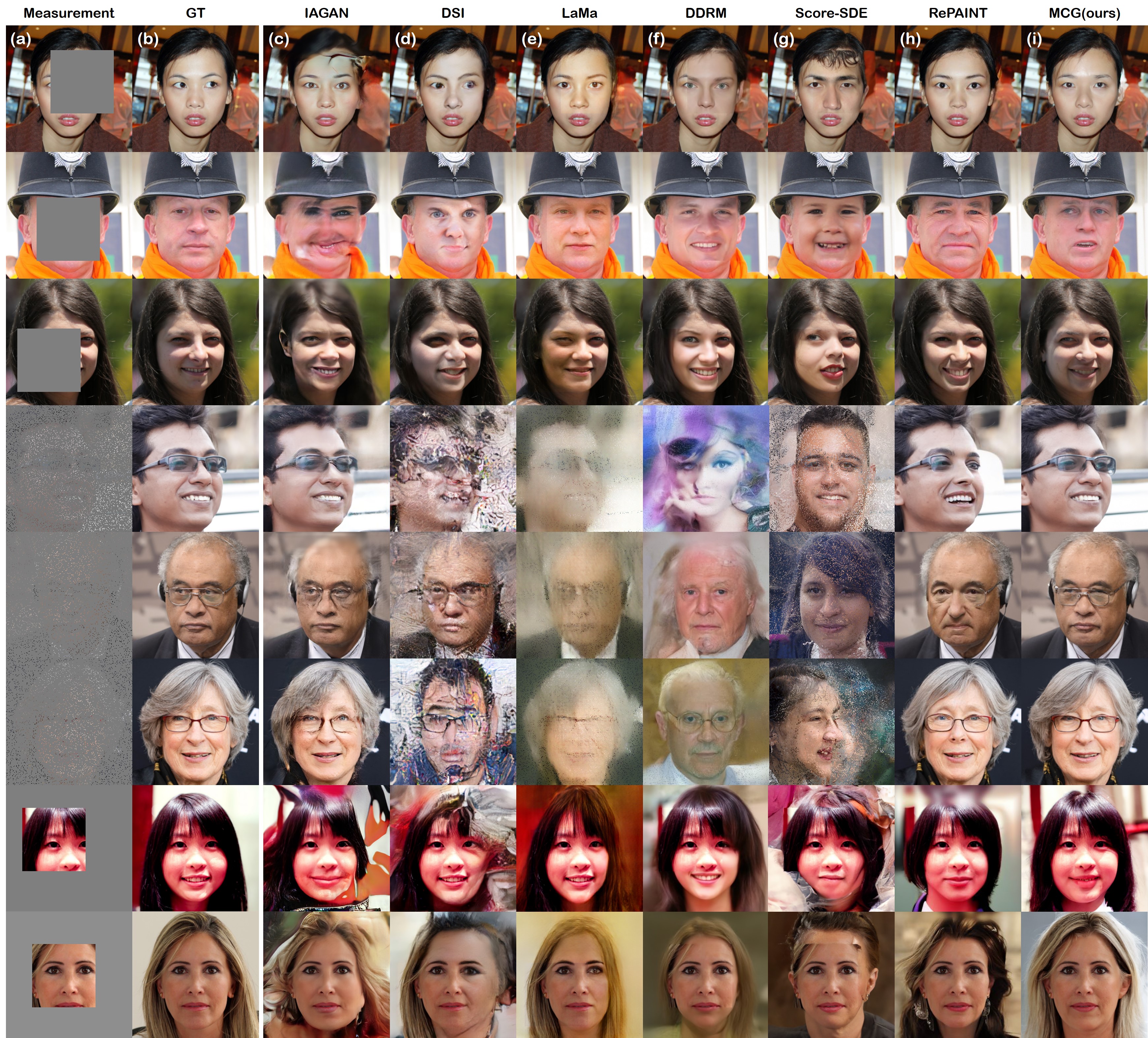}
    \caption{Inpainting results on FFHQ 256$\times$256 data. (a) Measurement, (b) ground truth, (c) IAGAN~\cite{hussein2020image}, (d) DSI~\cite{peng2021generating}, (e) LaMa~\cite{suvorov2022resolution}, (f) DDRM~\cite{kawar2022denoising}, (g) score-SDE~\cite{song2020score}, (h) RePAINT~\cite{lugmayr2022repaint}, (i) MCG (ours).} 
    \label{fig:results_inpainting_ffhq_all}
\end{figure}

\begin{figure}[t]
    \centering
    \includegraphics[width=1.0\textwidth]{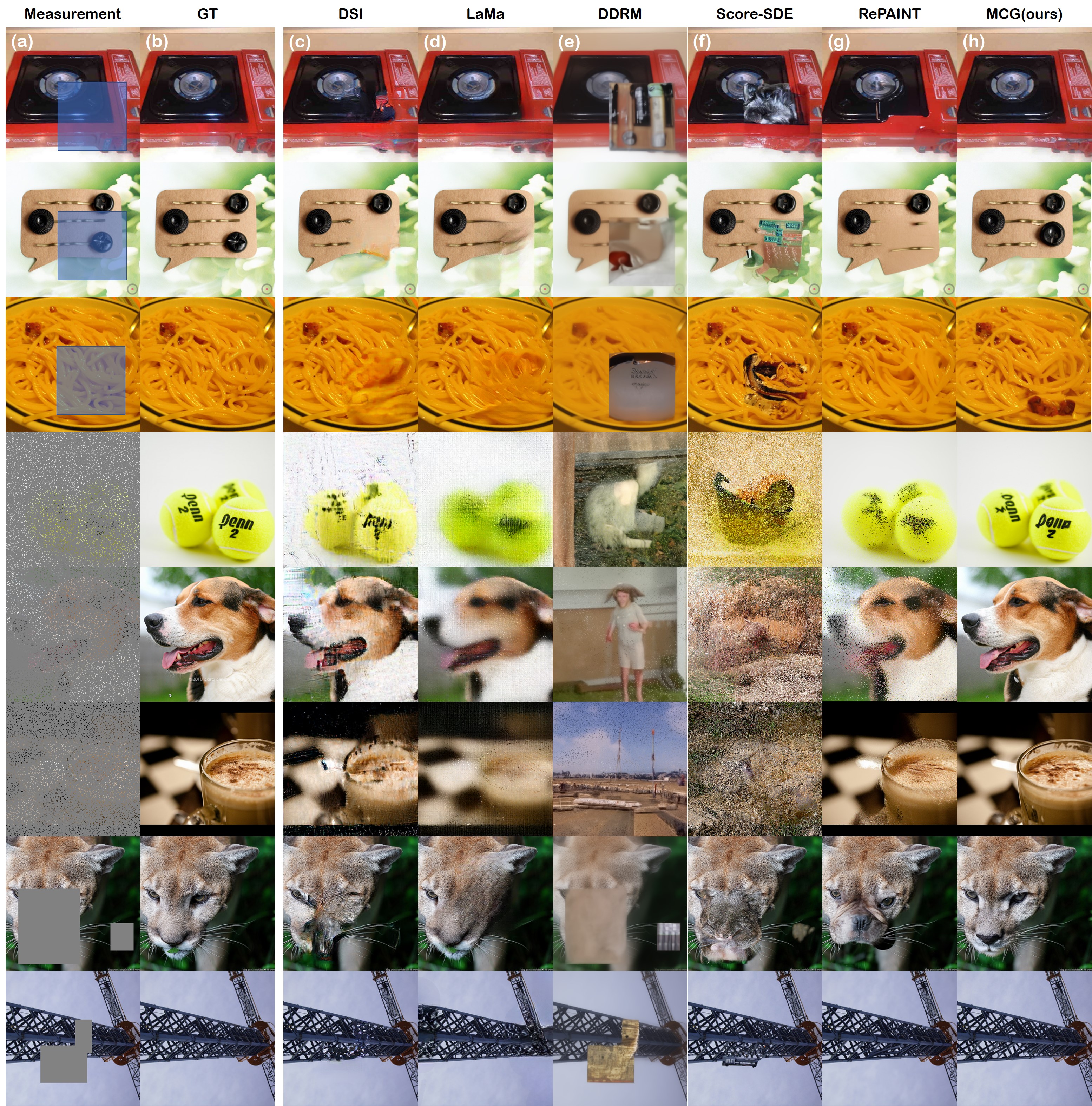}
    \caption{Inpainting results on ImageNet 256$\times$256 data.(a) Measurement, (b) ground truth, (c) DSI~\cite{peng2021generating}, (d) LaMa~\cite{suvorov2022resolution}, (e) DDRM~\cite{kawar2022denoising}, (f) score-SDE~\cite{song2020score}, (g) RePAINT~\cite{lugmayr2022repaint}, (h) MCG (ours).} 
    \label{fig:results_inpainting_imagenet_all}
\end{figure}

\begin{figure}[t]
    \centering
    \includegraphics[width=1.0\textwidth]{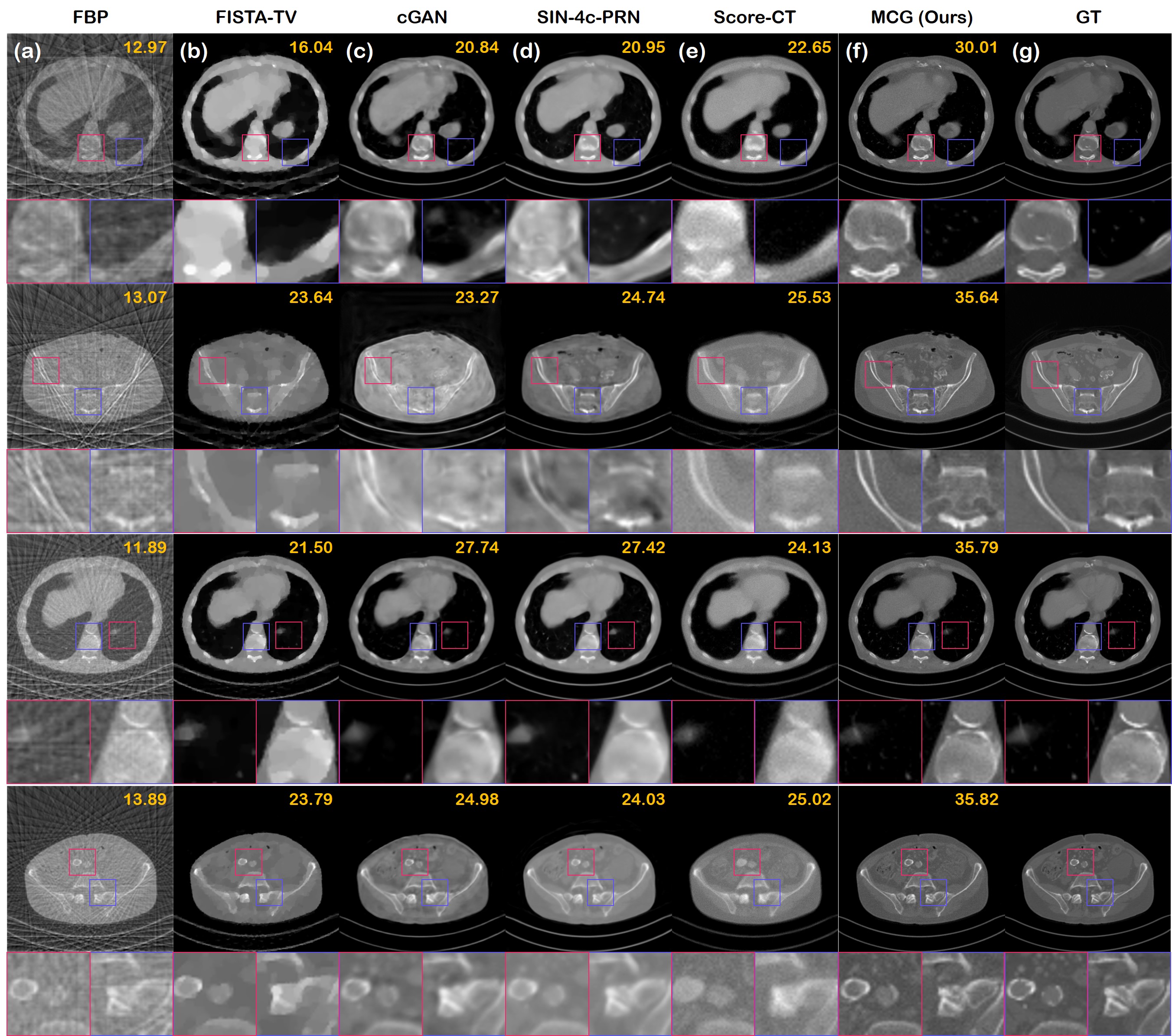}
    \caption{Sparse view CT reconstruction results on AAPM 256$\times$256 data.(a) FBP, (b) FISTA-TV~\cite{beck2009fast}, (c) cGAN~\cite{ghani2018deep}, (d) SIN-4c-PRN~\cite{wei20202}, (e) Score-CT~\cite{song2020score}, (f) MCG (Ours), (g) ground truth (GT).} 
    \label{fig:results_ct_all}
\end{figure}

\begin{figure}[t]
    \centering
    \includegraphics[width=1.0\textwidth]{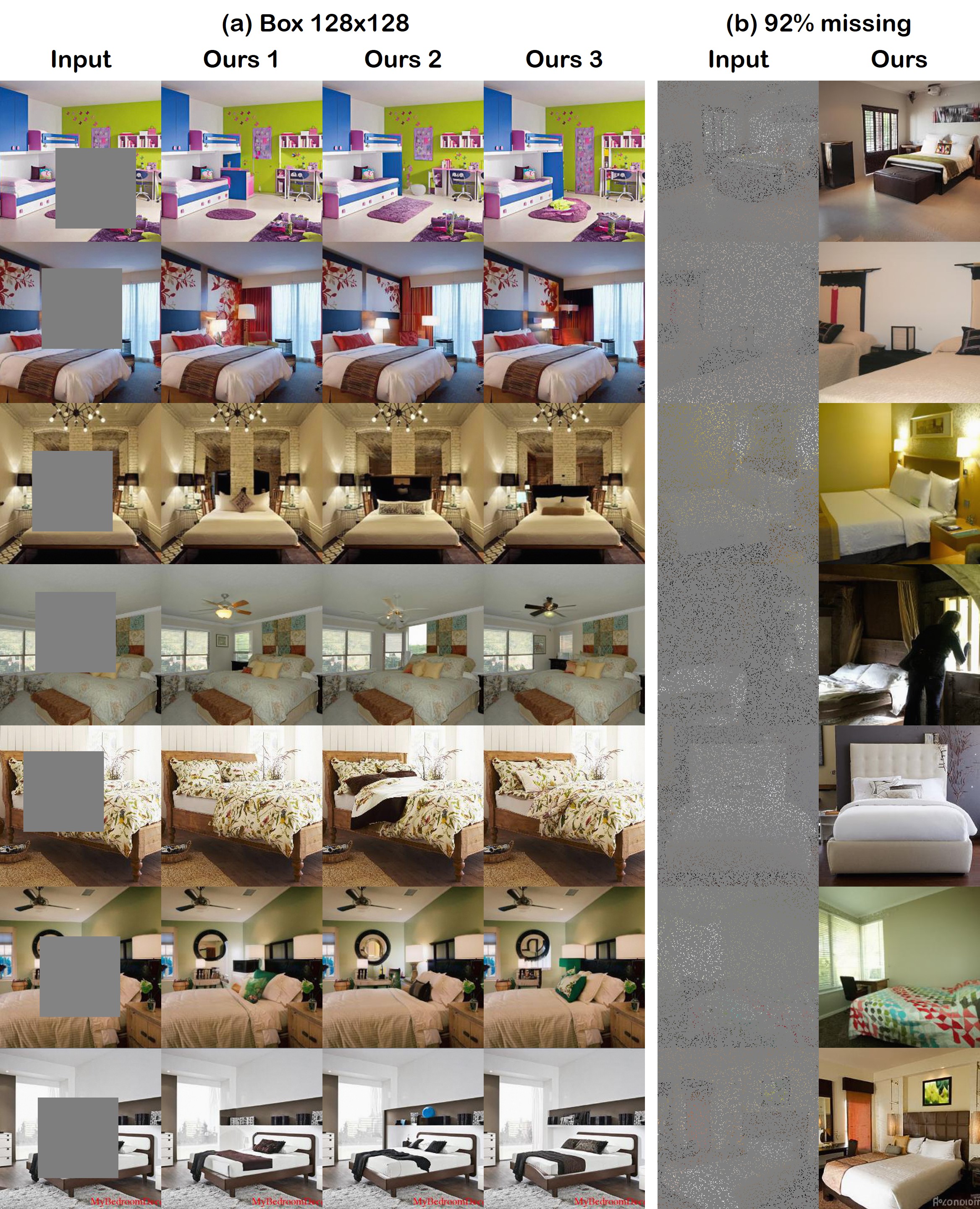}
    \caption{Inpainting results on LSUN-bedroom 256$\times$256 data with MCG. (a) Inpainting of 128$\times$128 box region. We show three stochastic samples generated with the proposed method. (b) 92 \% pixel missing imputation.} 
    \label{fig:results_inpainting_lsunbed_mult}
\end{figure}

\begin{figure}[t]
    \centering
    \includegraphics[width=1.0\textwidth]{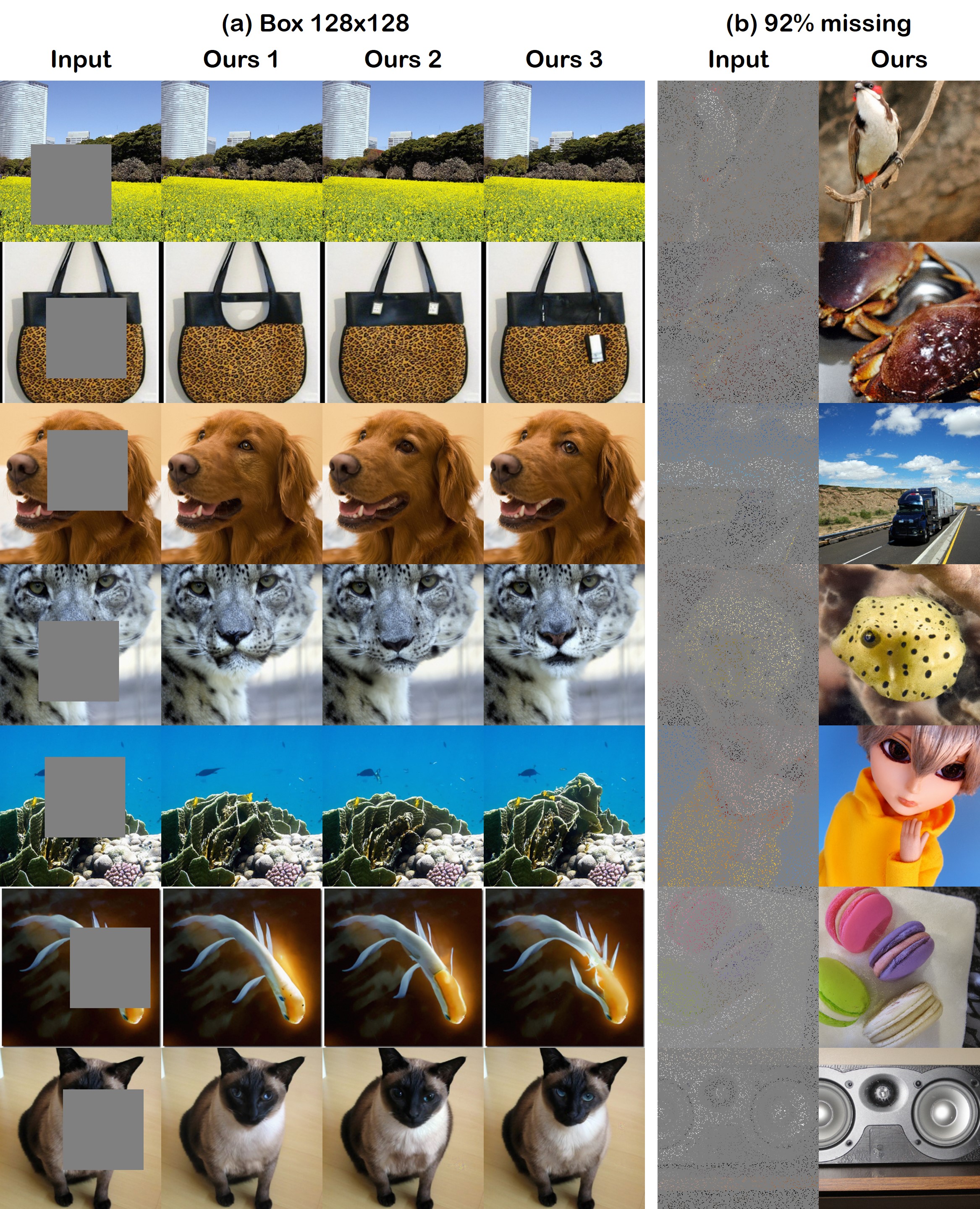}
    \caption{Inpainting results on ImageNet 256$\times$256 data with MCG. (a) Inpainting of 128$\times$128 box region. We show three stochastic samples generated with the proposed method. (b) 92 \% pixel missing imputation.} 
    \label{fig:results_inpainting_imagenet_mult}
\end{figure}

\begin{figure}[t]
    \centering
    \includegraphics[width=1.0\textwidth]{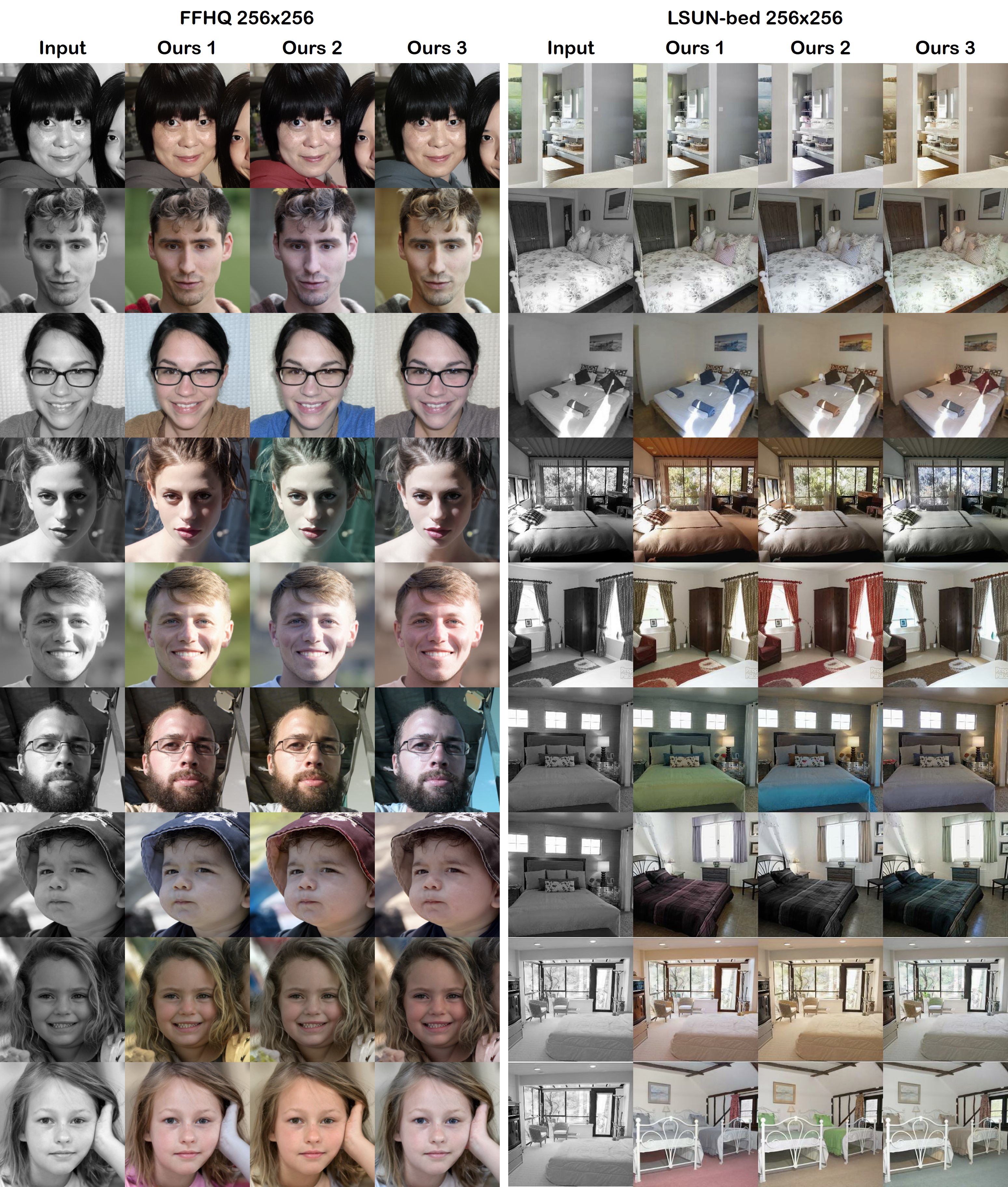}
    \caption{Colorization results on (left) FFHQ 256$\times$256 dataset, and (right) LSUN-bedroom 256$\times$256 dataset. We show 3 different reconstructions for each measurement that are sampled with the proposed method.} 
    \label{fig:results_color_mult}
\end{figure}

We provide extensive set of comparison study for each task in Fig.~\ref{fig:results_inpainting_ffhq_all},~\ref{fig:results_inpainting_imagenet_all}, and ~\ref{fig:results_ct_all}. Furthermore, in order to illustrate the ability of our method to generate multimodal reconstructions given a measurement, we present further experimental results of inpainting and colorization in the following figures: Fig.~\ref{fig:results_inpainting_ffhq_mult},~\ref{fig:results_inpainting_lsunbed_mult},~\ref{fig:results_inpainting_imagenet_mult}, and ~\ref{fig:results_color_mult}

\clearpage

\begin{figure}[t]
    \centering
    \includegraphics[width=1.0\textwidth]{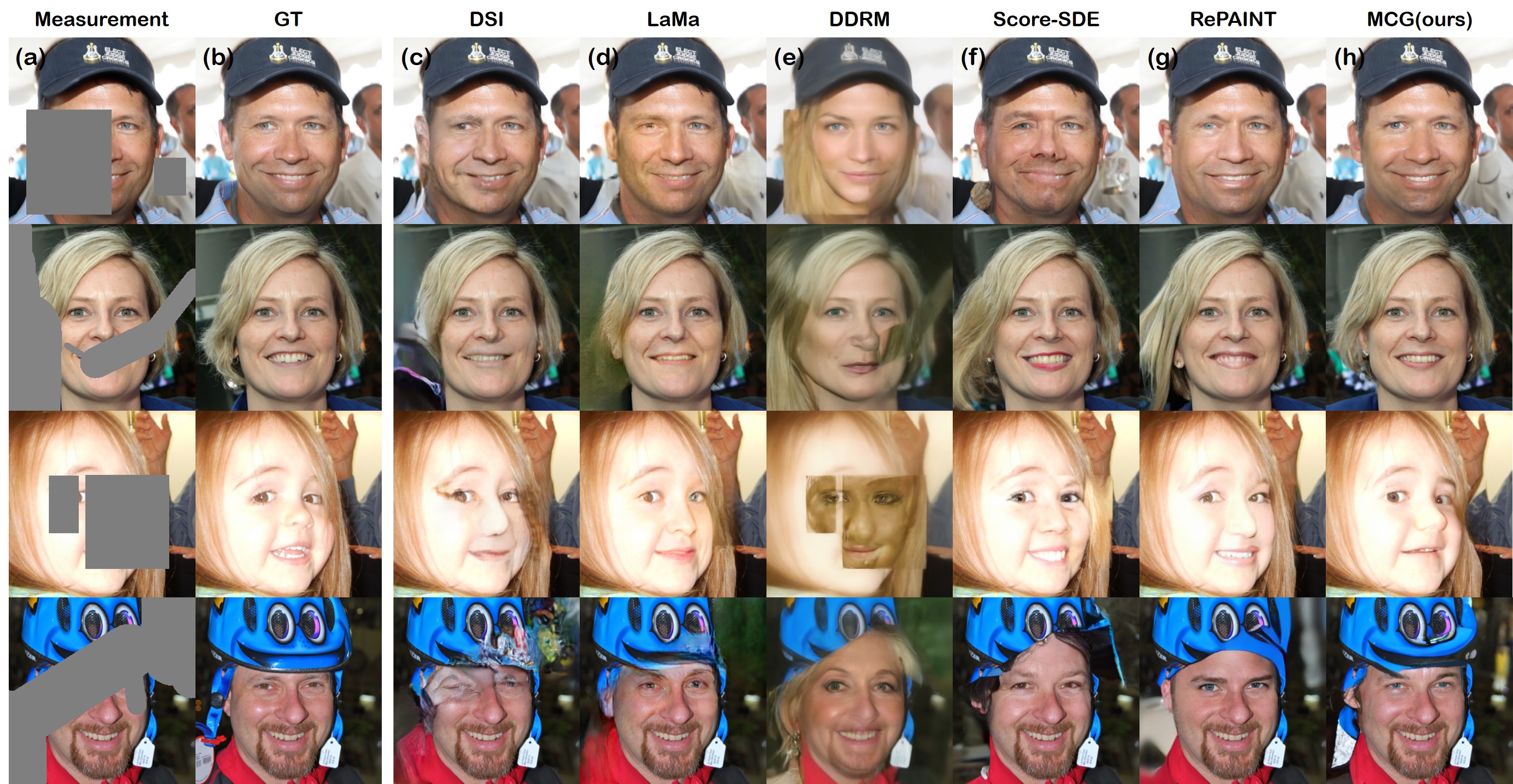}
    \caption{Inpainting results on FFHQ 256$\times$256 data with the LaMa~\cite{suvorov2022resolution} wide mask. (a) Measurement, (b) ground truth, (c) DSI~\cite{peng2021generating}, (d) LaMa~\cite{suvorov2022resolution}, (e) DDRM~\cite{kawar2022denoising}, (f) score-SDE~\cite{song2020score}, (g) RePAINT~\cite{lugmayr2022repaint}, (h) MCG (ours).} 
    \label{fig:results_inpainting_ffhq_lama_mask}
\end{figure}

\begin{figure}[t]
    \centering
    \includegraphics[width=1.0\textwidth]{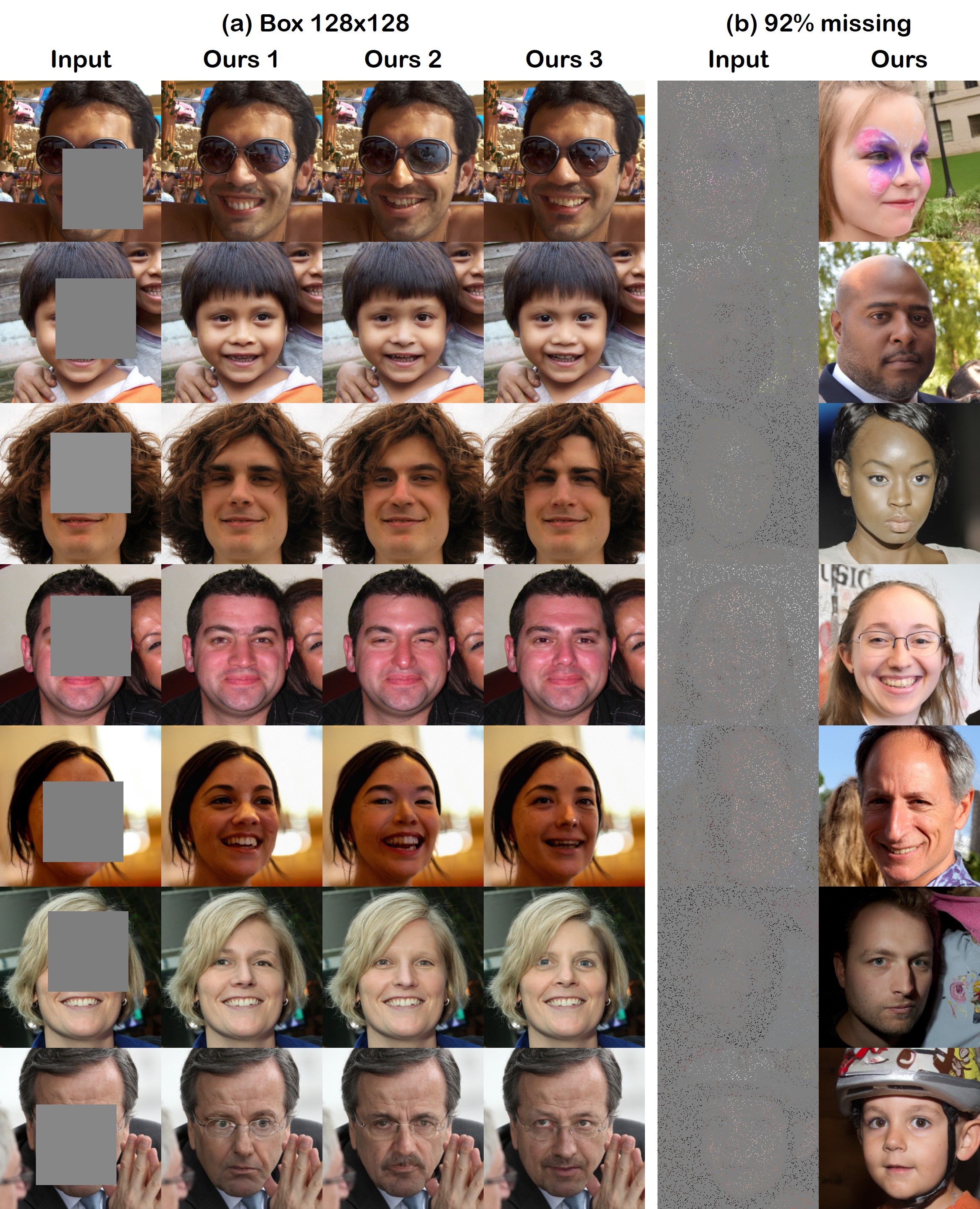}
    \caption{Inpainting results on FFHQ 256$\times$256 data with MCG. (a) Inpainting of 128$\times$128 box region. We show three stochastic samples generated with the proposed method. (b) 92 \% pixel missing imputation.} 
    \label{fig:results_inpainting_ffhq_mult}
\end{figure}

\end{document}

%% file: packages.tex
\usepackage[utf8]{inputenc} 
\usepackage[T1]{fontenc}    
\usepackage{hyperref}       
\usepackage{url}            
\usepackage{booktabs}       
\usepackage{amsfonts}       
\usepackage{nicefrac}       
\usepackage{microtype}      
\usepackage{xcolor}         
\usepackage{times}
\usepackage{epsfig}
\usepackage{graphicx}
\usepackage{placeins}
\usepackage{multirow}
\usepackage[skip=5pt]{caption}
\usepackage{rotating}
\usepackage{cancel}
\usepackage{setspace}
\usepackage{eso-pic}

\usepackage{bm}
\usepackage{bibunits} 

\usepackage{amssymb,amsthm}
\usepackage{hyperref}
\usepackage{amsmath}
\usepackage{graphicx}
\usepackage{wrapfig,lipsum,booktabs}

\usepackage{epsfig}
\usepackage{tikz}
\usetikzlibrary{spy}
\usepackage{algpseudocode}
\usepackage{algorithm}
\usepackage{mathrsfs}

\usepackage{nicefrac}       
\usepackage{booktabs}       

\usepackage{thmtools,thm-restate}
\usepackage{cleveref}

\usepackage{subcaption}        

%% file: macros.tex
\newcommand{\bb}{{\boldsymbol b}}

\newcommand{\x}{{\boldsymbol x}}
\newcommand{\s}{{\boldsymbol s}}
\newcommand{\y}{{\boldsymbol y}}
\newcommand{\z}{{\boldsymbol z}}
\newcommand{\ub}{{\boldsymbol u}}
\newcommand{\vb}{{\boldsymbol v}}
\newcommand{\f}{{\boldsymbol f}}
\newcommand{\w}{{\boldsymbol w}}
\newcommand{\Ib}{{\boldsymbol I}}
\newcommand{\Jb}{{\boldsymbol J}}
\newcommand{\Ab}{{\boldsymbol A}}
\newcommand{\Cb}{{\boldsymbol C}}

\newcommand{\Pb}{{\boldsymbol P}}
\newcommand{\Mb}{{\boldsymbol M}}
\newcommand{\Wb}{{\boldsymbol W}}
\newcommand{\Rb}{{\boldsymbol R}}
\newcommand{\Hb}{{\boldsymbol H}}
\newcommand{\Rd}{{\mathbb R}}
\newcommand{\Nd}{{\mathbb N}}
\newcommand{\Ed}{{\mathbb E}}
\newcommand{\Mc}{{\mathcal M}}

\newcommand{\Nc}{{\mathcal N}}

\newcommand{\Xc}{{\mathcal X}}

\DeclareMathOperator*{\argmin}{arg\,min}

\newtheorem{remark}{\bf{Remark}}

\definecolor{trolleygrey}{rgb}{0.5, 0.5, 0.5}
\definecolor{darkpastelgreen}{rgb}{0.01, 0.75, 0.24}
\definecolor{darkpink}{rgb}{0.91, 0.33, 0.5}
\definecolor{alizarin}{rgb}{0.82, 0.1, 0.26}
\definecolor{americanrose}{rgb}{1.0, 0.01, 0.24}

\makeatletter
\newcommand\footnoteref[1]{\protected@xdef\@thefnmark{\ref{#1}}\@footnotemark}
\makeatother

%% file: main.bbl
\begin{thebibliography}{10}

\bibitem{anderson1982reverse}
Brian~DO Anderson.
\newblock Reverse-time diffusion equation models.
\newblock {\em Stochastic Processes and their Applications}, 12(3):313--326,
  1982.

\bibitem{ardizzone2019guided}
Lynton Ardizzone, Carsten L{\"u}th, Jakob Kruse, Carsten Rother, and Ullrich
  K{\"o}the.
\newblock Guided image generation with conditional invertible neural networks.
\newblock {\em arXiv preprint arXiv:1907.02392}, 2019.

\bibitem{beck2009fast}
Amir Beck and Marc Teboulle.
\newblock A fast iterative shrinkage-thresholding algorithm for linear inverse
  problems.
\newblock {\em SIAM journal on imaging sciences}, 2(1):183--202, 2009.

\bibitem{bora2017compressed}
Ashish Bora, Ajil Jalal, Eric Price, and Alexandros~G Dimakis.
\newblock Compressed sensing using generative models.
\newblock In {\em International Conference on Machine Learning}, pages
  537--546. PMLR, 2017.

\bibitem{boyd2011distributed}
Stephen Boyd, Neal Parikh, Eric Chu, Borja Peleato, Jonathan Eckstein, et~al.
\newblock Distributed optimization and statistical learning via the alternating
  direction method of multipliers.
\newblock {\em Foundations and Trends{\textregistered} in Machine learning},
  3(1):1--122, 2011.

\bibitem{jax2018github}
James Bradbury, Roy Frostig, Peter Hawkins, Matthew~James Johnson, Chris Leary,
  Dougal Maclaurin, George Necula, Adam Paszke, Jake Vander{P}las, Skye
  Wanderman-{M}ilne, and Qiao Zhang.
\newblock {JAX}: composable transformations of {P}ython+{N}um{P}y programs,
  2018.

\bibitem{buzug2011computed}
Thorsten~M Buzug.
\newblock Computed tomography.
\newblock In {\em Springer handbook of medical technology}, pages 311--342.
  Springer, 2011.

\bibitem{choi2021ilvr}
Jooyoung Choi, Sungwon Kim, Yonghyun Jeong, Youngjune Gwon, and Sungroh Yoon.
\newblock {ILVR}: Conditioning method for denoising diffusion probabilistic
  models.
\newblock In {\em Proceedings of the IEEE/CVF International Conference on
  Computer Vision (ICCV)}, 2021.

\bibitem{chung2022come}
Hyungjin Chung, Byeongsu Sim, and Jong~Chul Ye.
\newblock {Come-Closer-Diffuse-Faster: Accelerating Conditional Diffusion
  Models for Inverse Problems through Stochastic Contraction}.
\newblock In {\em Proceedings of the IEEE/CVF Conference on Computer Vision and
  Pattern Recognition}, 2022.

\bibitem{daras2021intermediate}
Giannis Daras, Joseph Dean, Ajil Jalal, and Alexandros~G Dimakis.
\newblock Intermediate layer optimization for inverse problems using deep
  generative models.
\newblock In {\em International Conference on Machine Learning}, 2021.

\bibitem{de2020maximum}
Valentin De~Bortoli, Alain Durmus, Marcelo Pereyra, and Ana~Fernandez Vidal.
\newblock Maximum likelihood estimation of regularization parameters in
  high-dimensional inverse problems: an empirical bayesian approach. part ii:
  Theoretical analysis.
\newblock {\em SIAM Journal on Imaging Sciences}, 13(4):1990--2028, 2020.

\bibitem{deng2009imagenet}
Jia Deng, Wei Dong, Richard Socher, Li-Jia Li, Kai Li, and Li~Fei-Fei.
\newblock Imagenet: A large-scale hierarchical image database.
\newblock In {\em 2009 IEEE conference on computer vision and pattern
  recognition}, pages 248--255. Ieee, 2009.

\bibitem{dhariwal2021diffusion}
Prafulla Dhariwal and Alexander~Quinn Nichol.
\newblock Diffusion models beat {GAN}s on image synthesis.
\newblock In A.~Beygelzimer, Y.~Dauphin, P.~Liang, and J.~Wortman Vaughan,
  editors, {\em Advances in Neural Information Processing Systems}, 2021.

\bibitem{efron2011tweedie}
Bradley Efron.
\newblock Tweedie’s formula and selection bias.
\newblock {\em Journal of the American Statistical Association},
  106(496):1602--1614, 2011.

\bibitem{ghani2018deep}
Muhammad~Usman Ghani and W~Clem Karl.
\newblock Deep learning-based sinogram completion for low-dose ct.
\newblock In {\em 2018 IEEE 13th Image, Video, and Multidimensional Signal
  Processing Workshop (IVMSP)}, pages 1--5. IEEE, 2018.

\bibitem{gordon1970algebraic}
Richard Gordon, Robert Bender, and Gabor~T Herman.
\newblock Algebraic reconstruction techniques (art) for three-dimensional
  electron microscopy and x-ray photography.
\newblock {\em Journal of theoretical Biology}, 29(3):471--481, 1970.

\bibitem{NIPS2017_8a1d6947}
Martin Heusel, Hubert Ramsauer, Thomas Unterthiner, Bernhard Nessler, and Sepp
  Hochreiter.
\newblock Gans trained by a two time-scale update rule converge to a local nash
  equilibrium.
\newblock In I.~Guyon, U.~Von Luxburg, S.~Bengio, H.~Wallach, R.~Fergus,
  S.~Vishwanathan, and R.~Garnett, editors, {\em Advances in Neural Information
  Processing Systems}, volume~30. Curran Associates, Inc., 2017.

\bibitem{ho2020denoising}
Jonathan Ho, Ajay Jain, and Pieter Abbeel.
\newblock Denoising diffusion probabilistic models.
\newblock In {\em Advances in Neural Information Processing Systems},
  volume~33, pages 6840--6851, 2020.

\bibitem{ho2022video}
Jonathan Ho, Tim Salimans, Alexey Gritsenko, William Chan, Mohammad Norouzi,
  and David~J Fleet.
\newblock Video diffusion models.
\newblock {\em arXiv preprint arXiv:2204.03458}, 2022.

\bibitem{hussein2020image}
Shady~Abu Hussein, Tom Tirer, and Raja Giryes.
\newblock Image-adaptive gan based reconstruction.
\newblock In {\em Proceedings of the AAAI Conference on Artificial
  Intelligence}, volume~34, pages 3121--3129, 2020.

\bibitem{isola2017image}
Phillip Isola, Jun-Yan Zhu, Tinghui Zhou, and Alexei~A Efros.
\newblock Image-to-image translation with conditional adversarial networks.
\newblock In {\em Proceedings of the IEEE conference on computer vision and
  pattern recognition}, pages 1125--1134, 2017.

\bibitem{kadkhodaie2020solving}
Zahra Kadkhodaie and Eero Simoncelli.
\newblock Stochastic solutions for linear inverse problems using the prior
  implicit in a denoiser.
\newblock In {\em Advances in Neural Information Processing Systems},
  volume~34, pages 13242--13254. Curran Associates, Inc., 2021.

\bibitem{kang2017deep}
Eunhee Kang, Junhong Min, and Jong~Chul Ye.
\newblock A deep convolutional neural network using directional wavelets for
  low-dose x-ray ct reconstruction.
\newblock {\em Medical physics}, 44(10):e360--e375, 2017.

\bibitem{karras2019style}
Tero Karras, Samuli Laine, and Timo Aila.
\newblock A style-based generator architecture for generative adversarial
  networks.
\newblock In {\em Proceedings of the IEEE/CVF conference on computer vision and
  pattern recognition}, pages 4401--4410, 2019.

\bibitem{kawar2022denoising}
Bahjat Kawar, Michael Elad, Stefano Ermon, and Jiaming Song.
\newblock Denoising diffusion restoration models.
\newblock In {\em ICLR Workshop on Deep Generative Models for Highly Structured
  Data}, 2022.

\bibitem{kawar2021snips}
Bahjat Kawar, Gregory Vaksman, and Michael Elad.
\newblock Snips: Solving noisy inverse problems stochastically.
\newblock {\em Advances in Neural Information Processing Systems},
  34:21757--21769, 2021.

\bibitem{kazantsev2019ccpi}
Daniil Kazantsev, Edoardo Pasca, Martin~J Turner, and Philip~J Withers.
\newblock Ccpi-regularisation toolkit for computed tomographic image
  reconstruction with proximal splitting algorithms.
\newblock {\em SoftwareX}, 9:317--323, 2019.

\bibitem{kim2021noisescore}
Kwanyoung Kim and Jong~Chul Ye.
\newblock Noise2score: Tweedie{\textquoteright}s approach to self-supervised
  image denoising without clean images.
\newblock In A.~Beygelzimer, Y.~Dauphin, P.~Liang, and J.~Wortman Vaughan,
  editors, {\em Advances in Neural Information Processing Systems}, 2021.

\bibitem{kingma2013auto}
Diederik~P. Kingma and Max Welling.
\newblock Auto-encoding variational bayes.
\newblock In {\em 2nd International Conference on Learning Representations,
  {ICLR}}, 2014.

\bibitem{laumont2022bayesian}
R{\'e}mi Laumont, Valentin~De Bortoli, Andr{\'e}s Almansa, Julie Delon, Alain
  Durmus, and Marcelo Pereyra.
\newblock Bayesian imaging using plug \& play priors: when langevin meets
  tweedie.
\newblock {\em SIAM Journal on Imaging Sciences}, 15(2):701--737, 2022.

\bibitem{laurent2000adaptive}
Beatrice Laurent and Pascal Massart.
\newblock Adaptive estimation of a quadratic functional by model selection.
\newblock {\em Annals of Statistics}, pages 1302--1338, 2000.

\bibitem{lugmayr2022repaint}
Andreas Lugmayr, Martin Danelljan, Andres Romero, Fisher Yu, Radu Timofte, and
  Luc Van~Gool.
\newblock {RePaint: Inpainting using Denoising Diffusion Probabilistic Models}.
\newblock {\em arXiv preprint arXiv:2201.09865}, 2022.

\bibitem{mao2017least}
Xudong Mao, Qing Li, Haoran Xie, Raymond~YK Lau, Zhen Wang, and Stephen
  Paul~Smolley.
\newblock Least squares generative adversarial networks.
\newblock In {\em Proceedings of the IEEE international conference on computer
  vision}, pages 2794--2802, 2017.

\bibitem{ong2019local}
Frank Ong, Peyman Milanfar, and Pascal Getreuer.
\newblock Local kernels that approximate bayesian regularization and proximal
  operators.
\newblock {\em IEEE Transactions on Image Processing}, 28(6):3007--3019, 2019.

\bibitem{peng2021generating}
Jialun Peng, Dong Liu, Songcen Xu, and Houqiang Li.
\newblock {Generating diverse structure for image inpainting with hierarchical
  VQ-VAE}.
\newblock In {\em Proceedings of the IEEE/CVF Conference on Computer Vision and
  Pattern Recognition}, pages 10775--10784, 2021.

\bibitem{razavi2019generating}
Ali Razavi, Aaron Van~den Oord, and Oriol Vinyals.
\newblock Generating diverse high-fidelity images with vq-vae-2.
\newblock {\em Advances in neural information processing systems}, 32, 2019.

\bibitem{robbins1992empirical}
Herbert~E Robbins.
\newblock An empirical bayes approach to statistics.
\newblock In {\em Breakthroughs in statistics}, pages 388--394. Springer, 1992.

\bibitem{sarkka2019applied}
Simo S{\"a}rkk{\"a} and Arno Solin.
\newblock {\em Applied stochastic differential equations}, volume~10.
\newblock Cambridge University Press, 2019.

\bibitem{simonyan2014very}
Karen Simonyan and Andrew Zisserman.
\newblock Very deep convolutional networks for large-scale image recognition.
\newblock In {\em 3rd International Conference on Learning Representations,
  {ICLR}}, 2015.

\bibitem{song2022solving}
Yang Song, Liyue Shen, Lei Xing, and Stefano Ermon.
\newblock Solving inverse problems in medical imaging with score-based
  generative models.
\newblock In {\em International Conference on Learning Representations}, 2022.

\bibitem{song2020score}
Yang Song, Jascha Sohl{-}Dickstein, Diederik~P. Kingma, Abhishek Kumar, Stefano
  Ermon, and Ben Poole.
\newblock Score-based generative modeling through stochastic differential
  equations.
\newblock In {\em 9th International Conference on Learning Representations,
  {ICLR}}, 2021.

\bibitem{stein1981estimation}
Charles~M Stein.
\newblock Estimation of the mean of a multivariate normal distribution.
\newblock {\em The annals of Statistics}, pages 1135--1151, 1981.

\bibitem{suvorov2022resolution}
Roman Suvorov, Elizaveta Logacheva, Anton Mashikhin, Anastasia Remizova,
  Arsenii Ashukha, Aleksei Silvestrov, Naejin Kong, Harshith Goka, Kiwoong
  Park, and Victor Lempitsky.
\newblock Resolution-robust large mask inpainting with fourier convolutions.
\newblock In {\em Proceedings of the IEEE/CVF Winter Conference on Applications
  of Computer Vision}, pages 2149--2159, 2022.

\bibitem{tirer2018image}
Tom Tirer and Raja Giryes.
\newblock Image restoration by iterative denoising and backward projections.
\newblock {\em IEEE Transactions on Image Processing}, 28(3):1220--1234, 2018.

\bibitem{tooldkazanc}
Github~PK Tool, Nov Sun Mon Tue~Wed Thu, and Fri Sat.
\newblock dkazanc/tomobar.

\bibitem{van2017neural}
Aaron Van Den~Oord, Oriol Vinyals, et~al.
\newblock Neural discrete representation learning.
\newblock {\em Advances in neural information processing systems}, 30, 2017.

\bibitem{venkatakrishnan2013plug}
Singanallur~V Venkatakrishnan, Charles~A Bouman, and Brendt Wohlberg.
\newblock Plug-and-play priors for model based reconstruction.
\newblock In {\em 2013 IEEE Global Conference on Signal and Information
  Processing}, pages 945--948. IEEE, 2013.

\bibitem{vidal2020maximum}
Ana~Fernandez Vidal, Valentin De~Bortoli, Marcelo Pereyra, and Alain Durmus.
\newblock Maximum likelihood estimation of regularization parameters in
  high-dimensional inverse problems: An empirical bayesian approach part i:
  Methodology and experiments.
\newblock {\em SIAM Journal on Imaging Sciences}, 13(4):1945--1989, 2020.

\bibitem{wan2021high}
Ziyu Wan, Jingbo Zhang, Dongdong Chen, and Jing Liao.
\newblock High-fidelity pluralistic image completion with transformers.
\newblock In {\em Proceedings of the IEEE/CVF International Conference on
  Computer Vision}, pages 4692--4701, 2021.

\bibitem{wei20202}
Haoyu Wei, Florian Schiffers, Tobias W{\"u}rfl, Daming Shen, Daniel Kim,
  Aggelos~K Katsaggelos, and Oliver Cossairt.
\newblock 2-step sparse-view ct reconstruction with a domain-specific
  perceptual network.
\newblock {\em arXiv preprint arXiv:2012.04743}, 2020.

\bibitem{yu2015lsun}
Fisher Yu, Ari Seff, Yinda Zhang, Shuran Song, Thomas Funkhouser, and Jianxiong
  Xiao.
\newblock Lsun: Construction of a large-scale image dataset using deep learning
  with humans in the loop.
\newblock {\em arXiv preprint arXiv:1506.03365}, 2015.

\bibitem{zeng2022aggregated}
Yanhong Zeng, Jianlong Fu, Hongyang Chao, and Baining Guo.
\newblock Aggregated contextual transformations for high-resolution image
  inpainting.
\newblock {\em IEEE Transactions on Visualization and Computer Graphics}, 2022.

\bibitem{zhang2017learning}
Kai Zhang, Wangmeng Zuo, Shuhang Gu, and Lei Zhang.
\newblock Learning deep cnn denoiser prior for image restoration.
\newblock In {\em Proceedings of the IEEE conference on computer vision and
  pattern recognition}, pages 3929--3938, 2017.

\bibitem{zhang2018unreasonable}
Richard Zhang, Phillip Isola, Alexei~A Efros, Eli Shechtman, and Oliver Wang.
\newblock The unreasonable effectiveness of deep features as a perceptual
  metric.
\newblock In {\em Proceedings of the IEEE conference on computer vision and
  pattern recognition}, pages 586--595, 2018.

\bibitem{zhu2017unpaired}
Jun-Yan Zhu, Taesung Park, Phillip Isola, and Alexei~A Efros.
\newblock Unpaired image-to-image translation using cycle-consistent
  adversarial networks.
\newblock In {\em Proceedings of the IEEE international conference on computer
  vision}, pages 2223--2232, 2017.

\end{thebibliography}
